%% file: abstract.tex
\documentclass[twoside,11pt]{article}
\usepackage{jair, theapa, rawfonts}

\usepackage{graphicx}%
\usepackage{multirow}%
\usepackage{url} 
\usepackage{amsmath,amssymb,amsfonts}%
\usepackage{amsthm}%
\usepackage{mathrsfs}%
\usepackage[title]{appendix}%
\usepackage{xcolor}%
\usepackage{textcomp}%
\usepackage{manyfoot}%
\usepackage{booktabs}%
\usepackage{algorithm}%
\usepackage{algorithmicx}%
\usepackage{algpseudocode}%
\usepackage{listings}%

\usepackage{rotating}
\usepackage{svg}
\usepackage{amsmath}
\usepackage{hhline}
\usepackage{tikz,pgfplots,pgfplotstable}
\usepackage{caption} 
\usepackage{subcaption}
\usepackage{graphicx}
\usepackage{multicol}
\usepackage{tablefootnote}
\usepackage{forest}
\usepackage{pdflscape}
\usepackage{rotating}

\usepackage{makecell} %for new line within a cell
\usepackage{float}
\floatstyle{plaintop}
\restylefloat{table}

\makeatletter
\newcommand\footnoteref[1]{\protected@xdef\@thefnmark{\ref{#1}}\@footnotemark}
\makeatother

\pgfplotsset{compat=default}
\usetikzlibrary{pgfplots.groupplots}

\pgfplotsset{compat=1.11,
    /pgfplots/ybar legend/.style={
    /pgfplots/legend image code/.code={%
       \draw[##1,/tikz/.cd,yshift=-0.25em]
        (0cm,0cm) rectangle (3pt,0.8em);},
   },
}

\pgfkeys{/pgf/number format/.cd,
    fixed,
    set thousands separator={}}

\usepackage{xcolor}

\newcommand{\etal}{ \textit{et al.} }

\usepackage{longtable}
\usepackage{placeins} %for placing the table before references

\usepackage{tikz,pgfplots,pgfplotstable}
\definecolor{newcolor}{rgb}{.8,.349,.1}
\definecolor{bblue}{HTML}{4F81BD}
\definecolor{rred}{HTML}{C0504D}
\definecolor{ggreen}{HTML}{9BBB59}
\definecolor{ppurple}{HTML}{9F4C7C}
\definecolor{coolblack}{rgb}{0.0, 0.18, 0.39}
\definecolor{darkcyan}{rgb}{0.0, 0.55, 0.55}

\definecolor{q1}{HTML}{2f4f4f}
\definecolor{q2}{HTML}{008000}
\definecolor{q3}{HTML}{b8860b}
\definecolor{q4}{HTML}{ff4500}
\definecolor{q5}{HTML}{6a5acd}
\definecolor{q6}{HTML}{ffff00}
\definecolor{q7}{HTML}{00ff00}
\definecolor{q8}{HTML}{00bfff}
\definecolor{q9}{HTML}{0000ff}
\definecolor{q10}{HTML}{ff00ff}
\definecolor{q11}{HTML}{ff1493}
\definecolor{q12}{HTML}{7fffd4}
\definecolor{q13}{HTML}{ffb6c1}
\jairheading{81}{2024}{43-69}{06/2024}{09/2024}
\ShortHeadings{The State of Computer Vision Research in Africa}
{Omotayo\etal}
\firstpageno{1}

\begin{document}

\title{The State of Computer Vision Research in Africa} 

\author{\name Abdul-Hakeem Omotayo$^\dagger$ \email aaomotayo@ucdavis.edu \\ 
       \addr University of California, Davis, USA 
       \AND
       \name Ashery Mbilinyi $^\dagger$ \email ashery.mbilinyi@ubc.ca\\ 
       \addr University of British Columbia, Canada 
       \AND
       \name Lukman E. Ismaila $^\dagger$ \email lismail1@jhu.edu\\ 
       (Corresponding author) \addr Johns Hopkins University, USA 
       \AND
       \name Houcemeddine Turki \email turkiabdelwaheb@hotmail.fr \\
       \addr University of Sfax, Tunisia 
       \AND
       \name Mahmoud Abdien, Karim Gamal \email 21mmah,21kgmm@queensu.ca\\
       \addr Queen’s University, Canada
       \AND
       \name Idriss Tondji, Yvan Pimi \email itondji,ypimi@aimsammi.org\\
       \addr African Institute for Mathematical Sciences, Senegal
       \AND
       \name Naome A. Etori \email etori001@umn.edu\\
       \addr University of Minnesota-Twin Cities, USA
       \AND
       \name Marwa M. Matar \email marwamatar@azhar.edu.eg\\
       \addr Al-Azhar University in Cairo, Egypt
       \AND
       \name Clifford Broni-Bediako \email clifford.broni-bediako@riken.jp\\
       \addr RIKEN Center for Advanced Intelligence Project, Japan
       \AND
       \name Abigail Oppong \email abigoppong@gmail.com\\ 
       \addr Ashesi University, Ghana 
       \AND
       \name Mai Gamal \email mai.tharwat@guc.edu.eg\\
       \addr German University in Cairo, Egypt
       \AND
       \name Eman Ehab \email e.ehab@nu.edu.eg\\
       \addr Nile University, Egypt
       \AND
       \name Gbetondji Dovonon \email gbetondji.dovonon.22@ucl.ac.uk\\
       \addr University College London, UK
       \AND
       \name Zainab Akinjobi \email zakinjobi@ucdavis.edu\\
       \addr University of California, Davis, USA
       \AND
       \name Daniel Ajisafe \email dajisafe@cs.ubc.ca\\ 
       \addr University of British Columbia, Canada
       \AND
       \name Oluwabukola G. Adegboro \email oluwabukola.adegboro2@mail.dcu.ie\\
       \addr Dublin City University, Ireland
       \AND
       \name Mennatullah Siam \email mennatullah.siam@ontariotechu.ca\\
       \addr Ontario Tech University \\ \& University of British Columbia, Canada
       \AND
       \name 
       \addr $\dagger$ These~authors~contributed~equally~to~this~work\\
       %\addr Corresponding~author(s)~E-mail(s):~$lismail@jhu.edu$ 
       }

% For research notes, remove the comment character in the line below.
% \researchnote

\maketitle
\setcounter{page}{43}
\pagebreak
\begin{abstract}
Despite significant efforts to democratize artificial intelligence (AI), computer vision which is a sub-field of AI, still lags in Africa. A significant factor to this, is the limited access to computing resources, datasets, and collaborations. As a result, Africa's contribution to top-tier publications in this field has only been 0.06\% over the past decade. Towards improving the computer vision field and making it more accessible and inclusive, this study analyzes 63,000 Scopus-indexed computer vision publications from Africa. We utilize large language models to automatically parse their abstracts, to identify and categorize topics and datasets. This resulted in listing more than 100 African datasets. Our objective is to provide a comprehensive taxonomy of dataset categories to facilitate better understanding and utilization of these resources. We also analyze collaboration trends of researchers within and outside the continent. Additionally, we conduct a large-scale questionnaire among African computer vision researchers to identify the structural barriers they believe require urgent attention. In conclusion, our study offers a comprehensive overview of the current state of computer vision research in Africa, to empower marginalized communities to participate in the design and development of computer vision systems. %We propose strategies to advance the field, including improving access to datasets, fostering collaboration among African researchers and institutions, mainly focusing on topics aligned with the needs of African communities, and implementing capacity-building programs.
\end{abstract}

\section{Introduction}
\label{sec:intro}

\input{content/intro.tex}

\section{Related Works}
\label{sec:related}
\input{content/related}
\section{Methods}
\label{sec:method}
\input{content/method}

\section{African Computer Vision Datasets}
\label{sec:dataset}
\input{content/datasets}

\section{African Computer Vision Topics}
\input{content/topics}

\section{Publishing and Collaboration Trends}
\label{sec:geotemporal_analysis}
\input{content/publishing_trends}

\section{Large-scale Questionnaire}
\label{sec:survey}
\input{content/questionnaire}

\section{Conclusion}
\label{sec:conclusion}
\input{content/conclusion}

\section*{Acknowledgments}
\label{sec:acknowledgement}
\input{content/acknowledgement}

% \bibliographystyle{sn-apacite}
% \bibliography{refs}

% \section{Introduction}
% \label{Introduction}

% \section{Experimental Results}

% \label{results}

%  [section ommitted]

% \section{A Theoretical Model}
% \label{analysis}

%  [section ommitted]

% \section{Discussion}

%  [section ommitted]

% \acks{The authors wish to thank Hans-Martin Adorf, Don Rosenthal, 
% Richard Franier, Peter Cheeseman and Monte Zweben for their assistance
% and advice.  We also thank Ron Musick and our anonymous reviewers for
% their comments.  The Space Telescope Science Institute is operated by
% the Association of Universities for Research in Astronomy for NASA.
% }

% \appendix
% \section*{Appendix A. Probability Distributions for N-Queens}

% [section ommitted]

\vskip 0.2in
\bibliography{sample}
\bibliographystyle{theapa}

\end{document}

%% file: content/intro.tex
The field of computer vision allows machines to interpret visual data from images or videos, enabling various specialized tasks such as image classification, object recognition, image captioning, and scene understanding, among others. Recently, advancements in this field have led to the deployment of computer vision systems in various applications that directly impact society. These applications include remote sensing~(\shortciteA{sefala2021constructing,sirko2021continental}), medical image processing ~(\shortciteA{manescu2020weakly,nakasi2020web,roshanitabrizi2022ensembled}), and robotics (\shortciteA{grauman2022ego4d}). However, emerging studies have documented biases in some existing artificial intelligence methods,~\shortciteA{mehrabi2021survey,obermeyer2019dissecting,sweeney2013discrimination}, including computer vision~(\shortciteA{buolamwini2018gender,kinyanjui2020fairness}). These biases often originate from the lack of consultation with historically marginalized populations or their limited participation in the design of such methods, training datasets, or computer vision systems~(\shortciteA{mohamed2020decolonial}).

While various efforts have been made to empower marginalized communities in artificial intelligence, primarily driven by grassroots initiatives~(\shortciteA{fourie2023a}), there remains a significant gap in the field of computer vision, particularly in Africa. To address this gap, this work, as an initiative by \emph{Ro'ya}\footnote{\url{https://ro-ya-cv4africa.github.io/homepage/}}, a grassroots community focused on empowering African computer vision researchers, aims to bridge this divide. We have mainly focused on the African continent as a case study to investigate the state of computer vision research among marginalized populations. We specifically examine the topics being researched, the available datasets, and the structural barriers researchers face within the continent. Our aim is to decolonize computer vision research and empower marginalized communities by highlighting the challenges they encounter and the opportunities available~(\shortciteA{lewis2020indigenous,mhlambi2020rationality,mohamed2020decolonial,etori2024double,bondi2021envisioning}). Additionally, our work catalogs African computer vision datasets, making them more accessible. We provide a taxonomy of these datasets and analyze the regional distribution of research topics to offer a comprehensive overview of the current state of computer vision research in Africa. This analysis can help direct future research efforts and ensure they align with the needs of African communities. Furthermore, we emphasize the human aspect of the research process by conducting a large-scale questionnaire to understand the structural barriers faced by African researchers.

Our work is inspired by a previous research conducting a bibliometric study of African research in the context of machine learning for health (\shortciteA{turki2023machine}). However, unlike conventional bibliometric studies, we focus on documenting datasets, research topics, and researchers' view of the field, in addition to the conventional discussions on publishing trends. This work is an extension of our recent work~\shortciteA{omotayo2023towards}, that discussed publishing trends within African and global contexts in computer vision research.

%\textcolor{blue}{Our work is inspired by previous research that conducted a bibliometric study of African research in the context of machine learning for health~(\shortciteA{turki2023machine}). However, unlike conventional bibliometric studies, we focus on documenting datasets, research topics, and researchers’ views of the field, in addition to the conventional discussions on publishing trends. This work is an extension of our recent study~(\shortciteA{omotayo2023towards}) that discussed publishing trends within African and global contexts in computer vision research.}

%=================================================================================================

To summarize our contributions and insights: 
\begin{itemize}
\item We have gathered 96 official and 33 unofficial computer vision datasets, creating a taxonomy of African datasets organized into 31 categories based on applications and tasks involved. Our results show that forests, plants, and agriculture-related applications are the most prominent, while image classification is the top researched task. Overall, the taxonomy enhances the understanding and use of African datasets (Section \ref{sec:dataset}).

% We provide a taxonomy of African computer vision datasets, as the first endeavour to perform this, automated through large language models parsing our collected publications' abstracts (Section \ref{sec:dataset}). 
\item We have highlighted the top computer vision research topics in Africa and their regional distribution, demonstrating the diverse focus areas and geographical spread of research efforts across the continent. Our results show that topics like image segmentation are more prominent in Northern Africa, while research related to galaxy morphology is more prominent in Southern Africa. On the other hand, photogrammetry and remote sensing research is prominent in other regions (Section \ref{sec:topics}).

% We present a taxonomy of the top research topics and their distribution per African region in computer vision (Section \ref{sec:topics}).

%\item We present a pioneer bibliometric study on African computer vision, which
\item We document the inequities in computer vision research across the continent and the disparities in publishing venues. African research constitutes only 0.06\% of the total publication-researcher pairs in top-tier venues. Northern and Southern Africa are the two highest regions publishing in computer vision overall, comprising 88.5\% of the total publications. Based on these patterns, we suggest a pan-African approach to strengthen the overall continent’s research ecosystem (Section \ref{sec:geotemporal_analysis}).

% \item We document inequity in computer vision research in the continent and the disparities in publishing venues, then analyze the temporal trends and collaboration patterns. (Section \ref{sec:geotemporal_analysis}). 

\item Based on a large-scale questionnaire, we have examined regional disparities and identified common barriers and urgent priorities to enhance the African research ecosystem (Section \ref{sec:survey}).

% A large-scale quantitative survey (i.e., questionnaire) is conducted among computer vision and deep learning researchers in Africa through the Deep Learning Indaba platform (Section \ref{sec:survey}).
\end{itemize}

% \men{Mention EAAMO and that this is extension + citation}

% \men{Discussion of African computer vision datasets (official + unofficial) + Large-scale questionnaire}

% \men{Discuss Dataset organisation}

%% file: content/related.tex
%\men{Note: Review for modification}

In this section, we provide our closest related works, where our survey is focused on topics, datasets and researchers in addition to publishing trends. From the topics and publishing trends perspective, we found that bibliometric studies are the closest. From the datasets perspective, we cover other efforts on cataloging datasets with a regional focus, as in our case Africa. From the researchers perspective, we discuss different grassroots and organizations that encourage researchers from marginalized communities. While there has been previous questionnaires focused on African researchers~(\shortciteA{gaillard2001questionnaire,mbondji2014overview}), none discussed computer vision research which is our current focus.

\textbf{Bibliometric Studies.} In related scientometric and bibliometric studies, there exists multiple works that have studied scientific publications generally from Africa~(\shortciteA{pouris2014research,sooryamoorthy2021science}) or concerning a specific topic~(\shortciteA{tlili2022we,guleid2021bibliometric,musa2022bibliometric}) such as health sciences~(\shortciteA{musa2022bibliometric}) or COVID~(\shortciteA{guleid2021bibliometric}). One of the earliest studies~(\shortciteA{pouris2014research}) showed that African countries mostly focus on international collaborations rather than collaborating within the continent. They highlight that these are mainly driven by the availability of resources and interests outside Africa. 
%We think that encouraging these international collaborations is still an important component in improving African research. However, it is more important to empower the African research ecosystem to have some form of independence and increase the collaborations within the continent, while being connected to international collaborators. This is especially evident in computer vision, which is largely driven nowadays by large-scale datasets and models that are trained on expensive computing resources. Thus, we encourage international collaborations without undermining independent African research that focuses on our problems and needs. 

The recent work~(\shortciteA{turki2023machine}) studied African publications in machine learning for health. Their main findings indicated that Northern African countries had the most substantial contributions when compared to other African regions. However, this trend reduced over the years with more contributions emerging from sub-Saharan Africa. It also confirmed the correlation between international funding and collaborations in increasing the contributions from Africa. Inspired by this past work, we conduct a survey of African contributions, but we rather focus on the computer vision field which is more diverse with various applications from medical image processing to remote sensing. Some bibliometric studies focused on certain topics like convolutional neural networks~(\shortciteA{chen2020bibliometric}) or specific applications~(\shortciteA{iqbal2023last}). Yet, these do not focus on the African context which is our key question towards a decolonial computer vision approach that we encourage within our participatory framework. It is worth noting, that previous bibliometric studies focused on a high level analysis of the publications, e.g., identifying publishing and collaboration trends. In our case, we present another level of analysis relying on large language models parsing publications abstracts to automatically catalog African computer vision datasets.

%%This proves that the funding programs provided for African countries, the capacity building events like Deep Learning Indaba, and the free online and offline courses and mentorships in ML and biomedical informatics have succeeded to bridge the gap between African countries caused by financial burdens~\cite{vernon2019robotics}.

%Which scientific disciplines are emphasised in Africa?
%• How did research collaboration evolve in Africa during the period 2007–2011?
%• Who are the main research partners of African countries?
%• Are the patterns of collaboration (extended and disciplinary) in Africa similar to those in the rest of the world?
%• How do the various African countries perform in terms of collaboration?
%• Which are the main African institutions that are actively engaged in collaboration?

%bibliometric analysis on Africa~\cite{}. Generally Africa~\cite{pouris2014research,}.
\textbf{African Datasets.} As far as our knowledge extends, no previous studies provide a comprehensive and centralized platform that gathers African datasets and highlight the current state of computer vision research in Africa at large. We believe that currently, data science competition platforms like Kaggle\footnote{\url{https://www.kaggle.com}} and Zindi\footnote{\url{https://zindi.africa}} are the primary source of most of the unofficial African datasets. However, these platforms host data science competitions on various topics, including topics with or without African contexts. We are inspired by LANFRICA\footnote{\url{https://lanfrica.com/}}, an African platform that focuses on providing accessible African language datasets and natural language processing publications related to African languages. In this study, we focus primarily on computer vision datasets to facilitate easy access to them. One approach to listing African computer vision datasets is to start from a pre-defined taxonomy and gather dataset publications related to these predefined categories. We call this a top-down approach and we find it misleading, as it might not capture the African research landscape because of its dependence on the predefined categories. We rather choose a bottom-up approach that gathers all African computer vision publications in the last decade, followed by the classification of dataset papers using large language models and performing manual annotation of the categories of these datasets. To the best of our knowledge, we are the first to propose such a bottom-up approach to catalog computer vision datasets with a regional focus (i.e., Africa).

\textbf{Grass-roots Participatory Framework.} Recent work~(\shortciteA{bondi2021envisioning}) has discussed a critique of the definition of AI for social good and how to evaluate a project's goodness. They propose a PACT (People, Activities, Contexts, Technologies) framework; a participatory approach to enable capabilities in communities. In this framework, they provide a list of guiding questions that can help researchers assess the goodness in the project within a participatory framework. This relates to calls for a decolonial AI that was recently spread~(\shortciteA{birhane2020towards,kalluri2021don,lewis2020indigenous,mohamed2020decolonial,whittaker2019disability}), where one of these calls iterated on the importance of participatory and community based efforts~(\shortciteA{mohamed2020decolonial}). Towards a participatory grassroots framework~(\shortciteA{fourie2023a}), a discussion comparing top-down \textit{vs.} bottom-up approaches with a description of the types of grassroots communities that emerged recently were presented. These include: (i) affinity-based organizations such as Women in Machine Learning and Black in AI, (ii) topic-based communities such as \emph{Masakhane}, and \emph{SisonkeBiotik}, and (iii) event-based ones such as Deep Learning Indaba. The framework they proposed is focused on African grassroots, where they discuss the common values and participation roles within such communities. One specific community, SisonkeBiotik, is focusing on machine learning for health. Their initial project was a bibliometric study of African research in machine learning for health~(\shortciteA{turki2023machine}). Inspired by such efforts, we aim to perform a similar study focused on the computer vision field. 

However, unlike previous works, we rather provide it within a survey framework to catalog datasets, topics and researchers' view of the barriers existing in their field. Additionally, we use publications in top-tier venues and African contributions there to document quantitatively the inequity in research and assess Africans' access to opportunities in the field. We focus on that, since top tier publications could be an entry point for a lot of opportunities in terms of scholarships, research grants, and collaborations. It also can impact the kind of datasets and compute, African researchers have access to. 

%% file: content/method.tex
\begin{figure*}[t]
    \centering
       \includegraphics[width=\textwidth]{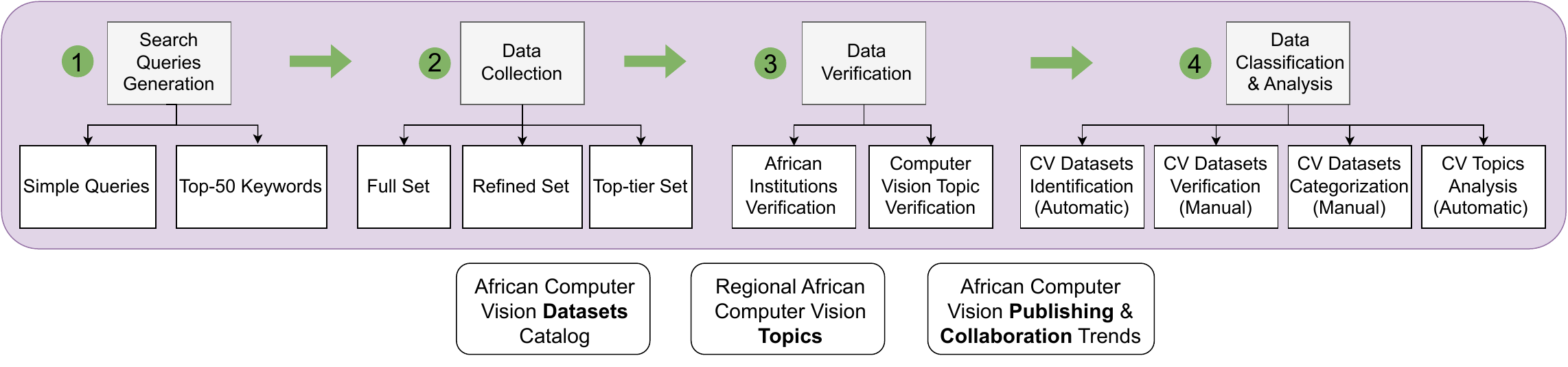}
    \caption{Our proposed pipeline for data collection, verification, and analysis of the African Scopus-indexed computer vision publications. The search query generation uses simple queries to retrieve all computer vision publications (i.e., \emph{full} set) or generates queries based on the top-50 keywords in computer vision as a sample (i.e., \emph{refined} set). This is followed by data collection of the \emph{full}, \emph{refined} and \emph{top-tier} publications sets and a verification phase on the \emph{refined} and \emph{top-tier} sets. Finally, we perform classification and analysis combining automatic, i.e., large language models parsing abstracts, and manual categorization.}
    \label{fig:methodfig}
\end{figure*}

In this section, we detail our method for gathering the necessary data for our survey with a focus on Scopus-indexed publications. Figure~\ref{fig:methodfig} describes four main stages: (i) automatic search query generation, (ii) data collection, (iii) data verification, and (iv) data classification and analysis. The computer vision field is quite broad and multidisciplinary. It overlaps with machine learning for health but also has a broader set of applications beyond this. We aim to collect three types of data. The first type is African publications that are Scopus-indexed with any relevance to computer vision in general, we refer to this set as the \emph{full} data. The second type is a reduced set of African publications that focuses on the top-50 keywords used in the computer vision field, we refer to this set as the \emph{refined} data. Finally, we collect the publications in top-tier venues in the computer vision field as the \emph{top-tier} set. In the following, we describe the details of the aforementioned four stages and the collection of these three sets of data. 

\subsection{Query Generation and Data Collection}
In the \emph{full} publications set, where we aim to collect all African publications relevant to the computer vision field, we restrict our search query to African countries and the search keyword to \emph{(``image'' OR ``computer vision'')}. Publications with at least one author from an African institution are the ones considered. We also restrict the time interval from 2012 to 2022 with the start of the deep learning era when convolutional neural networks won the 2012 ImageNet~(\shortciteA{deng2009imagenet}) challenge~(\shortciteA{krizhevsky2017imagenet}). The \emph{full} publications set is approximately 63,000 publications. Due to the large-scale nature of this set, we do not verify it but we use it to provide insights on the topics distribution per African region.

As for the \emph{refined} publications set, we only focus on the top-50 keywords used in computer vision publications to reduce our data and allow for a consecutive verification phase. This set includes approximately 18,000 publications where the search query was restricted to African countries similar to the \emph{full} set. The search keyword used is \emph{(``image'' OR ``computer vision'') AND (KEYWORD)}, where \emph{KEYWORD} was replaced with noun phrases from the top-50 computer vision keywords that include \emph{``deep learning'', ``object detection'', ``image segmentation'', ``robotics''} as examples. The top-50 keywords were retrieved using an off-the-shelf tool as described later. This refined and reduced set of publications allowed for a verification phase to remove false positives. False positives stem from being wrongly assigned as relevant to computer vision or having at least one African author. 

Finally, for the \emph{top-tier} publications set we focus on conferences and journals that are well-acknowledged in the computer vision field without any geographical restrictions. We use the CORE\footnote{\url{http://portal.core.edu.au/conf-ranks/}} system to identify rank $\text{A}^*, \text{A}$ computer vision venues and a few of the machine learning ones that include computer vision publications. The top-tier conferences we use include: \emph{CVPR, ICCV, ECCV, ICML, ICLR, NeurIPS} and we add \emph{MICCAI} for medical image processing publications. For the top-tier journals, we focus on \emph{TPAMI} and \emph{IJCV}. The final \emph{top-tier} set has approximately 45,000 publications. We do acknowledge that some of the publications in \emph{ICML, ICLR} and \emph{NeurIPS} can be on general machine learning, but we see it still as a good statistic for understanding how African institutions contribute to these venues. 
%Moreover, we provide a separate analysis of the most frequent and established computer vision venue (i.e., \emph{CVPR}) to avoid the aforementioned issue. 
We also perform a verification on this set, to remove false positives in terms of publication types such as ``Retracted'' or ``Review''.

Throughout the data collection stage for the \emph{refined} and \emph{top-tier} sets we use Scopus APIs\footnote{\url{https://pybliometrics.readthedocs.io/en/stable}}. We mildly use SciVal\footnote{\label{scival}\url{https://scival.com/home}} in the selection of the top-50 keywords and the \emph{full} set collection, as it has constrained usage. For example, it has limited control over the time-interval and the retrieved meta-data such as the authors' countries and affiliation history.

\subsection{Data Verification}
\label{sec:method_details}
We conduct a verification phase to reduce sources of errors to have a better understanding of African computer vision research. It is even more important when working on such a diverse topic as computer vision. The verification phase includes a combination of automatic and manual verification that we will describe in detail. Since the \emph{full} set is around 63,000 entries we find it difficult to verify and rather focus on the \emph{refined} and \emph{top-tier} sets. 

The reason we choose to perform an initial verification phase as we found three sources of errors in the \emph{refined} and \emph{top-tier} sets. The sources of errors include: (i) African authorship errors, (ii) computer vision relevance issues, and (iii) irrelevant publication types. We present some examples of these sources of errors and issues. For the African authorship errors, one example is labelling \emph{Papua New Guinea} as an African country conflating it with \emph{Guinea} which was identified in multiple publications. Others include typos in the country name such as replacing Switzerland with Swaziland. As for the computer vision relevance issues, one example is the use of the term "image data" in the abstract as an example but has no relevance to the topic being covered in the publication which operates on a simulated milling circuit dataset. We gather and show the aforementioned examples of publications that were rejected during our verification phase in our code repository~\footnote{\label{code}\url{https://github.com/Ro-ya-cv4Africa/acvsurvey/}}. We use a broad definition of relevance to computer vision. We consider works that operate on image datasets to interpret these images as relevant to computer vision. Finally, for the publication types we filter out "Retracted", "Review", "Erratum", or "Conference Review" types, to focus on the research publications. %\men{Reminder to check the aforementioned paper details to confirm 1D signal part.}

The verification phase for the computer vision relevance starts with automatic verification by filtering out venues with less than ten publications. We use a condition that the venue name does not belong to any of the top-tier venues defined previously and does not include the keywords \emph{image} or \emph{computer vision} in the venue name. These types of venues such as \emph{Solar Energy} are less relevant to the computer vision field. Although this might result in false negatives where some of the relevant publications might be filtered out, we favoured reduced false positives in the \emph{refined} set. 
%and we show ~(\shortciteA{Hamed2020310}) as an example
%Since the \emph{full} set covers all the data contrasting both sets during the analysis can help us in a better understanding of the research landscape. 
The rest of the venues we inspect them manually to identify potential venues that could be irrelevant to the computer vision topic. For these potential venues, we randomly sample publications from them to manually inspect their titles and abstracts to conclude whether these publications and venues are relevant to computer vision or not. 

Additionally, we verify publications' authors and affiliations to ensure that they include at least one African institution. We start with an automatic verification of the affiliations' countries. At that stage, \emph{Papua New Guinea} was identified as a false positive. This is followed by randomly sampling publications and manually inspecting the authors. 
Finally, we filter out "Retracted", "Editorial", "Review", "Erratum", "Conference Review" publication types to focus on research publications. The final \emph{refined} set after verification is around 12,000 publications. We retrieve the full metadata of the authors for both the \emph{refined} and \emph{top-tier} sets.

%We conduct three types of analysis on the collected datasets towards a better understanding of the African computer vision research which include: (i) A geo-temporal analysis where we take into consideration the five African regions and their publishing patterns over time. (ii) The collaboration patterns analysis where we contrast collaborations across African countries with respect to international collaborations. We also study whether African authors are key contributors as first or last authors or not. (iii) 

\subsection{Data Classification and Analysis}
%\men{Hakeem: can you write here how we did the classification of computer vision dataset papers and you can refer to the annotation guide we used as well here. Then follow this with how we did the topics analysis and others like publishing and collaborations trends. Thank you}

In this section, we describe a distinctive bottom-up approach that was employed for the identification and annotation of African publications that focus on computer vision datasets. This method diverged from the traditional top-down approach, which typically starts with predefined categories for data collection. The initial phase involved identifying potential dataset papers by using large language models (i.e., GPT series~(\shortciteA{brown2020language})) to analyze their abstracts. This helped us in gathering a comprehensive range of African computer vision dataset publications, which we categorized under official datasets. This extensive collection ensured a broad and inclusive base for the subsequent annotation process. Additionally, a collection of unofficial datasets gathered from challenge websites and data host platforms that were not formally published were manually gathered from the last five years.

Annotation and categorization were central to our bottom-up approach. Each dataset publication underwent a manual annotation process, where two descriptive labels were assigned by annotators. These labels could be selected from a predefined list or created independently by the annotators, allowing for a more flexible categorization. The predefined list is selected from the categories used in \emph{CVPR}~\footnote{\url{https://cvpr.thecvf.com/Conferences/2023/CallForPapers}} conference to define the different research topics. This list was then augmented with additional categories found by our annotators, hence the bottom-up approach. It provides a more comprehensive but still computer vision focused overview of the field in Africa. This is in contrast to the ACM computing classification system, which offers a top-down classification for the entire computing field. The annotators, were selected from a group of PhD students and postdocs in relevant fields; computer vision, medical image processing, and neuroscience. They were responsible for labeling the datasets based on their abstracts (for official publications) or descriptions (for unofficial ones). This process was iterative, with guidelines refined based on annotators' feedback after the initial ten entries, ensuring a continuously improving classification system. Each dataset publication was annotated by two separate annotators. When conflict occurs between these two annotators, a third annotator, as a senior researcher, resolves the conflict and provides the final labels for the dataset.

Ethical considerations were also a significant aspect of the methodology. Annotators were instructed to identify any datasets that posed ethical concerns, such as potential privacy violations or issues with the right to be forgotten. The approach was inclusive, allowing for the categorization of papers that did not directly propose new datasets but augmented existing ones. Additionally, datasets originating from African institutions but not exclusively about African subjects were included, with appropriate notes for clarification.

The outcome of this meticulous process was the development of a comprehensive taxonomy of African datasets in computer vision. This taxonomy, born from a flexible, inclusive, and ethically aware approach, provides a valuable resource for current research and lays a foundational framework for future explorations in the field. 

%\textcolor{blue}{The taxonomy of datasets was developed with a focus on computer vision tasks and application areas, inspired by Computer Vision and Pattern Recognition (CVPR) conference categories on their list of topics~\footnote{\url{https://cvpr.thecvf.com/Conferences/2023/CallForPapers}}. It is then augmented by new categories that were identified by the annotators as part of our bottom-up approach, which provides a more comprehensive overview of the field of computer vision in Africa. This is in contrast to the ACM Computing Classification System, which offers a top-down classification for the entire computing field.}

Moreover, we investigate the full publications set in addition to datasets identification and categorization, using three forms of analysis: (i) A topic analysis where we highlight what kind of computer vision research problems are mostly tackled in the African continent with regional distribution. For this analysis we rely on the keywords of each publication and we use an off-the-shelf tool, SciVal\footnoteref{scival}, that we believe is performing a standard procedure for clustering. The top three keywords for each publication is retrieved under ``Topic Name''. This analysis helps us to retrieve the top research topics in computer vision and the regional trends within these. (ii) A geo-temporal analysis where we take into consideration the five African regions and their publishing patterns over time. (iii) The collaboration patterns analysis where we contrast collaborations across African countries with respect to international collaborations.

%Beyond these three analyses, we also study whether African authors are key contributors as first or last authors or not.

%Furthermore, We conduct two additional types of analysis on the collected datasets towards a better understanding of the African computer vision research which include: (i) A geo-temporal analysis where we take into consideration the five African regions and their publishing patterns over time. (ii) The collaboration patterns analysis where we contrast collaborations across African countries with respect to international collaborations. We also study whether African authors are key contributors as first or last authors or not.

%that performs clustering on the index keywords for the publications, using the term-frequency and inverse document frequency. The final keywords are the most weighted node in each corresponding cluster. %Beyond conveying statistics and numbers we share stories from the community since one of the golden standard of participatory research is ``community first''. ..}

\subsection{Ethical Considerations}
%report here the ethical considerations of the dataset we have collected:
We report some ethical considerations for the collected dataset that can guide our future work. Since we only focus on Scopus-indexed publications, there exists a bias to the English language. It is worth noting that the French language is widely used in multiple African countries, but unfortunately, our dataset does not include these publications or the ones in African languages. Finally, we document that our research team includes researchers from Nigeria, Cameroon, Benin, Egypt, Tunisia, Tanzania and Ghana spanning four African regions (Western, Northern, Eastern and Central regions). Our team composition aligns with our goals to improve equity in research within Africa. 

\subsection{Manual Validation Results}

We verify the results of our \emph{refined} set, in terms of the topic, that it belongs to computer vision, and the authors, that they include at least one African author. We randomly select 5\% of the refined set of publications, that were manually verified by the team. Each paper was assigned two annotators. For the topic verification, we manually inspected the abstract, as for the authors we ensured at least one African institution existed in the affiliations. Our results for the topic verification show 91.1\% accuracy with only 8.9\% of the publications wrongly assigned as computer vision. In the authors' verification, we report 98.4\% accuracy, which demonstrates that most of our errors were from irrelevant topics rather than the authors. Moreover, we report high inter-annotator agreement of 93.9\% and 96.7\% in both the topic and the authors' verification respectively.
%\begin{itemize}
%\item scopus indexed is only English publications bias in Language
%\item Our way of using affiliation history has bias to authors that original come from Northern and Southern Africa since these are reportedly the ones mostly publishing from their institutions. Hence, other regions where institutions have less chance to publish will not be documented in the affiliation history and the authors and will be hard to retrieve. Future work ....
%\item stories conveyed in this paper are biased towards the community members themselves which necessitates for our future work ....
%\end{itemize}

%% file: content/datasets.tex
\input{graphs/datasets_taxonomy}
African computer vision datasets face several challenges despite efforts from the local communities to collect and promote high-quality data~\cite{okolo2023responsible}. To consider all contributions with African data, we explore unofficial datasets in addition to those that have been officially published. Our bottom-up approach resulted in 96 officially published datasets and 33 unofficial ones published in data hosting and data science competitions platforms. 

We show our taxonomy of datasets and corresponding number per category in Figure~\ref{fig:datasets_tax} for a total of 31 categories, which shows the variability of the topics with a strong emphasis on benefiting African communities. Our taxonomy categorizes the datasets with respect to applications (e.g., ``remote sensing'' and ``wildlife related'') and tasks (e.g., ``object detection'' and ``image segmentation''). We show under applications 19 categories, while tasks has 12 categories. When we look at certain application categories, e.g., ``forests, plants and agriculture related'', ``document analysis and understanding'',  or ``animals, wildlife related'' are all applications that have a positive impact on the continent. Looking at the top five categories, most datasets were on general computer vision tasks of ``image classification'' (60 datasets) and ``object detection'' (36 datasets). This was followed by datasets that were related to ``Humans face, body, pose, gestures or movement'' (22 datasets). Finally, on the applications side we have both ``forests, plants, agriculture related'' at 18 datasets and ``document analysis'' ones at 17 datasets.
\input{tables/datasets_official_notrotated}

\input{tables/datasets_unofficial_notrotated}

\input{tables/datasets_citations}
In Table~\ref{tab:official_dataset} and~\ref{tab:nonofficial_dataset}, we provide only 10 entries from our collection of African computer vision datasets for each, the officially published and the unofficial ones, respectively. In the officially published ones, the datasets included at least one author affiliated with an African institution. These datasets include Ego4D~(\shortciteA{grauman2022ego4d}) for egocentric videos which was collected from around 74 worldwide locations including Africa. It opens up opportunities for robotics and augmented reality applications that can be fostered through international collaborations. Another dataset, ZeroWaste~(\shortciteA{bashkirova2022zerowaste}), is for automatic waste detection. It is a challenging dataset for in-the-wild industrial-grade waste detection and segmentation which provides harder scenarios for the detection algorithms. While the data itself is collected in the United States, it still provides opportunities to develop algorithms for efficient waste management which can be useful in Africa. The Hausa Visual Genome~(\shortciteA{abdulmumin2022hausa}) is designed for multi-modal machine translation for English to Hausa with images caption, where the Hausa language is used in around eight African countries. The remaining datasets were quite diverse in topics including COVID prediction~(\shortciteA{muhammad2022deep}), malaria detection~(\shortciteA{manescu2020weakly}), UAV imagery for agricultural monitoring~(\shortciteA{amraoui2022avo}), scanned books for the Arabic language~(\shortciteA{elanwar2021extracting}), land cover mapping and change detection~(\shortciteA{lelong2022land,yuh2019effects}) and even datasets related to weather monitoring~(\shortciteA{hunt2020dataset}). All of which have various benefits to the African communities. Moreover, we show in Table~\ref{tab:datasets_citations} the top-5 cited datasets retrieved from both Scopus (left), Google Scholar (right) using pybliometrics and scholarly third party libraries.

\input{tables/funding_agencies}
For the full collection of African computer vision datasets refer to our publicly available dataset repository~\footnote{\url{https://github.com/Ro-ya-cv4Africa/acvdatasets}}. Additionally, we show the publicly available datasets and provide their access. Some of the official ones include datasets published in CVPR, IJDAR, LREC. Through our full data we found multiple African datasets publishing in \emph{Data in Brief}, although not a high impact journal, yet it provides a source for African data that can be used for small-scale projects and challenges in Africa. 

For the unofficial datasets, it was difficult to retrieve the year of publishing the challenge as it was not necessarily shared and dependant on the challenge platform. We mainly provide its availability and provide their access, in addition to the country where the data was collected. In the unofficial datasets (see Table~\ref{tab:nonofficial_dataset}) it is mostly retrieved from Zindi, the African data science competition platform. In these collected datasets, Kenya is among the top dataset contributors. 

Finally, we provide a list of some of the funding agencies that encourage African dataset acquisition (Table \ref{tab:funding_agencies}). Each agency or program has a specific aim, such as supporting innovation hubs, using technology and data for social justice, or promoting economic development. While there are multiple initiatives promoting data collection for research and development across the African continent, specific institutions provide funding for particular regions, notably Sub-Saharan Africa, to bridge their productivity gap (\shortciteA{turki2023machine}).

%% file: graphs/datasets_taxonomy.tex
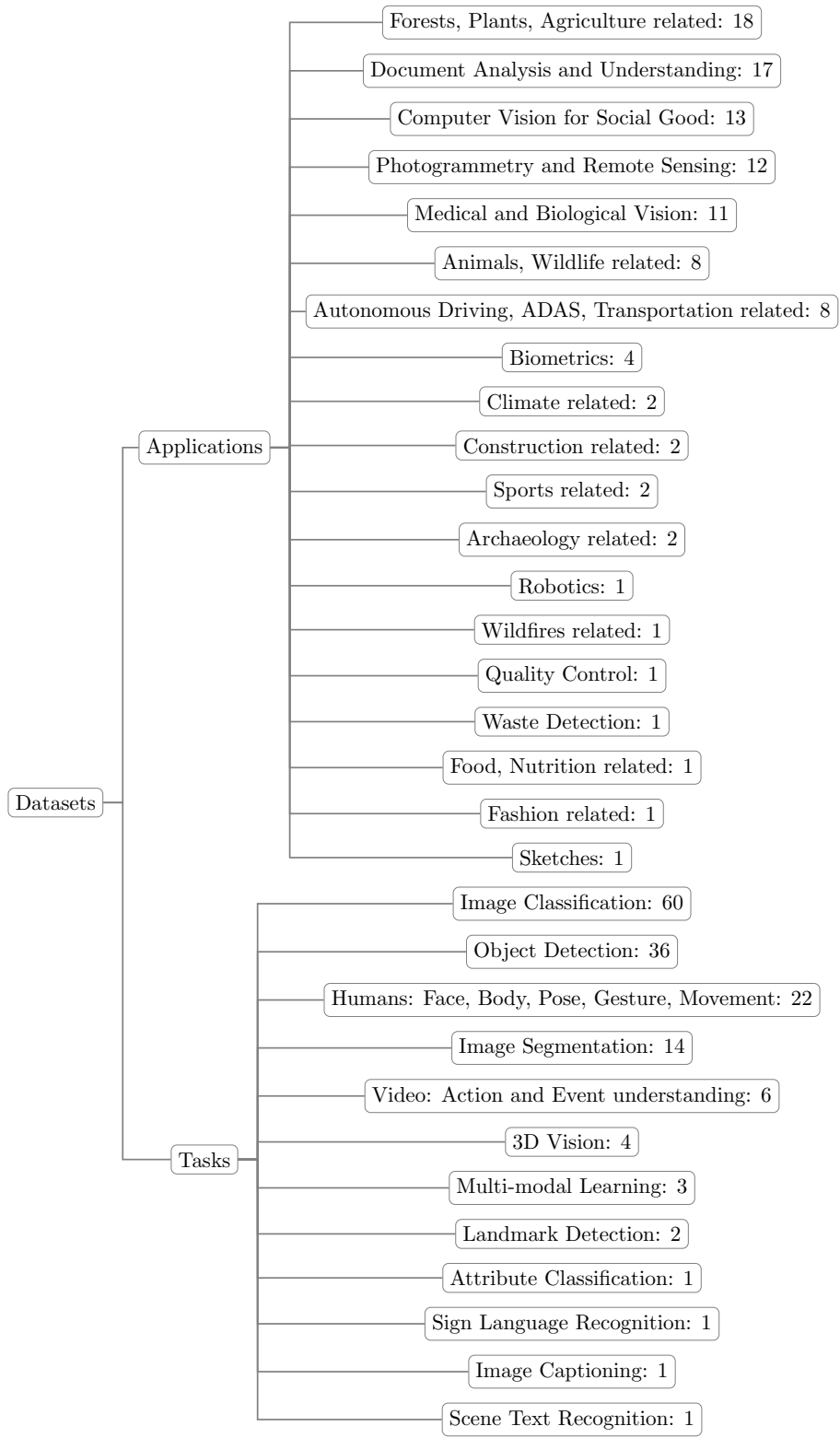
\begin{figure}[!htbp]
\resizebox{0.8\textwidth}{!}{
\centering
\begin{forest} for tree={%
    draw=gray, minimum width=1cm, minimum height=.5cm, rounded corners=3,
    l sep+=5pt,
    grow'=east,
    edge={gray, thick},
    parent anchor=east,
    child anchor=west,
    if n children=0{tier=last}{},
    edge path={
      \noexpand\path [draw, \forestoption{edge}] (!u.parent anchor) -- +(10pt,0) |- (.child anchor)\forestoption{edge label};
    },
    if={isodd(n_children())}{
      for children={
        if={equal(n,(n_children("!u")+1)/2)}{calign with current}{}
      }
    }{}
    }
[Datasets
    [Applications
        [{Forests, Plants, Agriculture related}: 18]
        [Document Analysis and Understanding: 17]
        [Computer Vision for Social Good: 13]
        [Photogrammetry and Remote Sensing: 12]
        [Medical and Biological Vision: 11]
        [{Animals, Wildlife related}: 8]
        [{Autonomous Driving, ADAS, Transportation related}: 8]
        [Biometrics: 4]
        [Climate related: 2]
        [Construction related: 2]
        [Sports related: 2]
        [Archaeology related: 2]
        [Robotics: 1]
        [Wildfires related: 1]
        [Quality Control: 1]
        [Waste Detection: 1]
        [{Food, Nutrition related}: 1]
        [Fashion related: 1]
        [Sketches: 1]
    ]
    [Tasks
        [Image Classification: 60]
        [Object Detection: 36]
        [{Humans: Face, Body, Pose, Gesture, Movement}: 22]
        [Image Segmentation: 14]
        [Video: Action and Event understanding: 6]
        [3D Vision: 4]
        [Multi-modal Learning: 3]
        [Landmark Detection: 2]
        [Attribute Classification: 1]
        [Sign Language Recognition: 1]
        [Image Captioning: 1]
        [Scene Text Recognition: 1]
    ]
]
\end{forest}
}
\caption{Taxonomy of the datasets categories for the retrieved officially published datasets and unofficial ones hosted in challenges and data host platforms. We show, ``category: number of datasets retrieved under this category''.}
\label{fig:datasets_tax}
\end{figure}

%% file: tables/datasets_official_notrotated.tex
\begin{table*}[ht]
%\begin{sidewaystable*}[!htbp]

{\fontsize{7}{9}\selectfont
\caption{Officially published datasets with at least one African institution.}
\label{tab:official_dataset}
\begin{tabular}{|c|l|l|c|l|}

\hhline{|=====|}
& \textbf{Category} & \textbf{Dataset Name} & \textbf{Available} & \textbf{Venue} \\
\hhline{|=====|}

1 & \makecell[l]{Video: Action and Event Understanding\\ \\ Humans: Face, Body, \\Pose, Gesture, Movement} & Ego4D Dataset~$(1)$ & \checkmark\footnotemark & CVPR \\
\hline
2 & \makecell[l]{Object Detection\\ \\ Waste Detection} & ZeroWaste Dataset~$(2)$ & \checkmark\footnotemark & CVPR \\
\hline
3 & \makecell[l]{Multi-modal Learning\\ \\Image Captioning } & Hausa Visual Genome~$(3)$ & \checkmark\footnotemark & LREC \\
\hline
4 & \makecell[l]{Medical and Biological Vision\\ \\Image Classification} & Predicting COVID-19 $(4)$ & \checkmark\footnotemark & \makecell[l]{Trends and \\Advancements \\of Image \\Processing \\and its \\Applications} \\
\hline
5 & \makecell[l]{Forests, Plants, Agriculture related\\ \\Image Segmentation} & Avo-AirDB $(5)$ &\checkmark\footnotemark & Data in Brief \\
\hline
6 & \makecell[l]{Photogrammetry and Remote Sensing\\ \\Image Segmentation} & \makecell[l]{Land cover map of Vavatenina\\ region (Madagascar)~$(6)$ }& \checkmark\footnotemark\checkmark\footnotemark & Data in Brief\\
\hline
7 & \makecell[l]{Document Analysis and Understanding\\ \\Image Classification} & Scanned Arabic Books~$(7)$ & \checkmark\footnotemark & IJDAR \\
\hline
8 & \makecell[l]{Medical and Biological Vision\\ \\Object Detection} & \makecell[l]{Malaria and Sickle Cells Detection\\ in Blood Films~$(8)$ }& \checkmark\footnotemark & MICCAI \\
\hline
9 & \makecell[l]{Climate related\\ \\Video: Action and Event Understanding} & \makecell[l]{Photographed Lightning Events\\ SA~$(9)$ } & \checkmark\footnotemark & Data in Brief \\
\hline
10 & \makecell[l]{Animals, Wildlife related\\ \\Photogrammetry and Remote Sensing} & \makecell[l]{Effects of Land cover change \\on Great Apes distribution\\ in South East Cameroon~$(10)$ } & - & Scientific Reports \\
\hline

\end{tabular}
}
%\caption{Example table with checkmarks and footnotes}
\tiny{$(1)$ \cite{grauman2022ego4d}\\
$(2)$ \cite{bashkirova2022zerowaste}\\
$(3)$ \cite{abdulmumin2022hausa}\\
$(4)$ \cite{muhammad2022deep}\\
$(5)$ \cite{amraoui2022avo} \\
$(6)$ \cite{lelong2022land}\\
$(7)$ \cite{elanwar2021extracting}\\
$(8)$ \cite{manescu2020weakly}\\
$(9)$ \cite{hunt2020dataset}\\
$(10)$ \cite{yuh2019effects}}
%\end{sidewaystable*}
\end{table*}

% Footnote texts
\addtocounter{footnote}{-10} % Adjust the counter to the number of footnotes
\stepcounter{footnote}\footnotetext{\url{https://ego4d-data.org}}
\stepcounter{footnote}\footnotetext{\url{http://ai.bu.edu/zerowaste/}}
\stepcounter{footnote}\footnotetext{\url{http://hdl.handle.net/11234/1-4749}}
\stepcounter{footnote}\footnotetext{\url{https://drive.google.com/file/d/1bum9Sehb3AzUMHLhBMuowPKyr\_PCrB3a/view}}
\stepcounter{footnote}\footnotetext{\url{https://data.mendeley.com/datasets/tvhh83r3hj/2}}
\stepcounter{footnote}\footnotetext{\url{https://doi.org/10.18167/DVN1/XJCXCS}}
\stepcounter{footnote}\footnotetext{\url{https://doi.org/10.18167/DVN1/KDNOV4}}
\stepcounter{footnote}\footnotetext{\url{https://github.com/wdqin/BE-Arabic-9K}}
\stepcounter{footnote}\footnotetext{\url{https://doi.org/10.5522/04/12407567}}
\stepcounter{footnote}\footnotetext{\url{http://dx.doi.org/10.17632/44vkdvrn67.1}}
%\stepcounter{footnote}\footnotetext{\url{http://apes.eva.mpg.de/}}

% ... more footnote texts ...

%% file: tables/datasets_unofficial_notrotated.tex
\begin{table*}[ht]
%\begin{sidewaystable*}[!htbp]
%\centering
{\fontsize{7}{9}\selectfont 
\caption{Unofficial African Datasets.} 
\label{tab:nonofficial_dataset}
\begin{tabular}{|c|l|l|c|l|c|}
\hhline{|======|}
   & \textbf{Category}      & \textbf{Dataset Name}  & \textbf{Available}  & \textbf{Short Description}  & \textbf{Country}     \\ 
\hhline{|======|}

1  & \makecell[l]{Image Classification\\ \\Autonomous Driving, ADAS,\\ Transportation related}  & Pothole & \checkmark\footnotemark & \makecell[l]{Images of South \\African streets with \\or without  potholes} & \makecell{South \\Africa}  \\
\hline
2  & \makecell[l]{Food, Nutrition related\\ \\Image Classification} & Nigeria Food AI Dataset & \checkmark\footnotemark & \makecell[l]{Images of 14 \\distinct indigenous \\Nigerian dishes } & Nigeria  \\ 
\hline
3 & \makecell[l]{Forests, Plants, Agriculture \\related\\ \\Image Classification}  & Kenya Crop Type Detection &  \checkmark\footnotemark     & \makecell[l]{Images of crop fields}  &  Kenya  \\ 
\hline
4 & \makecell[l]{Image Segmentation\\  \\Humans: Face, Body, \\Pose, Gesture, Movement} & Spot the Mask Challenge  & \checkmark\footnotemark & \makecell[l]{Images of people \\wearing masks.} & - \\ 
\hline
5  &\makecell[l]{Image Segmentation \\ \\Image Classification} & \makecell[l]{Road Segment Identification} &\checkmark\footnotemark  & \makecell[l]{Images of different\\ landscapes with or\\ without road segments} & \makecell[c]{South \\Africa }                                                          \\ 
\hline
6 & \makecell[l]{Animals, Wildlife related\\ \\Object Detection} & \makecell[l]{Local Ocean Conservation\\~Sea Turtle Face Detection} & \checkmark\footnotemark & Images of sea turtles. & Kenya \\                   
\hline
7 & \makecell[l]{Image Classification} & \makecell[l]{Bill Classification in\\Tunisia Challenge} & \checkmark\footnotemark & \makecell[l]{Images of receipts\\ in restaurants,\\ parking lots,\\ and others} & Tunisia \\ 
\hline
8  & \makecell[l]{Image Classification\\ \\Object detection}  & \makecell[l]{Computer Vision for License\\~Plate Recognition Challenge}& \checkmark\footnotemark  & \makecell[l]{Images of vehicle\\ licence plates}  & Tunisia \\
\hline
9 & \makecell[l]{Forests, Plants, Agriculture \\related\\ \\Object Detection} & \makecell[l]{Digital Africa Plantation \\Counting Challenge} & \checkmark\footnotemark  & \makecell[l]{Images containing\\ palm trees} & \makecell[c]{C$\hat{o}$te\\d’Ivoire} \\ 
\hline
10  & \makecell[l]{Image Classification\\ \\Humans: Face, Body, \\Pose, Gesture, Movement} &\makecell[l]{Task Mate Kenyan Sign\\Language Classification\\Challenge} & \checkmark\footnotemark  & \makecell[l]{Images containing\\ Kenyan sign\\ language gestures} & Kenya \\ 
\hline
\end{tabular}}
%\caption{Example table with checkmarks and footnotes}

%\end{sidewaystable*}
\end{table*}

% Footnote texts
\addtocounter{footnote}{-10} % Adjust the counter to the number of footnotes
\stepcounter{footnote}\footnotetext{\url{https://zindi.africa/competitions/miia-pothole-image-classification-challenge/data}} %1
\stepcounter{footnote}\footnotetext{\url{https://www.kaggle.com/datasets/elinteerie/nigeria-food-ai-dataset/data}} %2
\stepcounter{footnote}\footnotetext{\url{https://www.kaggle.com/datasets/warcoder/kenya-crop-type-detection}} %3
\stepcounter{footnote}\footnotetext{\url{https://zindi.africa/competitions/spot-the-mask/data}} %4
\stepcounter{footnote}\footnotetext{\url{https://zindi.africa/competitions/road-segment-identification/data}}%5
\stepcounter{footnote}\footnotetext{\url{https://zindi.africa/competitions/local-ocean-conservation-sea-turtle-face-detection/data}}
\stepcounter{footnote}\footnotetext{\url{https://zindi.africa/competitions/bill-classification-in-tunisia-challenge/data}}
\stepcounter{footnote}\footnotetext{\url{https://zindi.africa/competitions/ai-hack-tunisia-2-computer-vision-challenge-2/dat}}
\stepcounter{footnote}\footnotetext{\url{https://zindi.africa/competitions/digital-africa-plantation-counting-challenge/data}}
\stepcounter{footnote}\footnotetext{\url{https://zindi.africa/competitions/kenyan-sign-language-classification-challenge/data}}

% ... more footnote texts ...

%% file: tables/datasets_citations.tex
\begin{table}[t]
\centering
\begin{tabular}{|ll|ll|}
\hline
\multicolumn{2}{|c|}{Google Scholar}    & \multicolumn{2}{c|}{Scopus} \\
Dataset & Citations  & Dataset & Citations \\ \hline
Ego4D~$(1)$ & 699 & Ego4D~$(1)$ & 219 \\
11K Hands~$(2)$ & 179 & 11K Hands~$(2)$ & 89 \\
Oil Palm Ghana~$(3)$ & 111 & Oil Palm Ghana~$(3)$ & 74 \\
Traffic Signs Detection~$(4)$ & 93 & Lake Hawassa Watershed~$(6)$ & 72 \\
Arabic Digit Recognition~$(5)$ & 70 & Traffic Signs Detection~$(4)$ & 71 \\ \hline
\end{tabular}
\caption{Top-5 most cited datasets with at least one author from an African institution retrieved from both Google Scholar and Scopus. The citation count is retrieved at 2-8-2024 for context of the presented datasets.
$(1)$~\cite{grauman2022ego4d}
$(2)$~\cite{afifi201911k}
$(3)$~\cite{chemura2015determination}
$(4)$~\cite{ayachi2020traffic}
$(5)$~\cite{el2007two}
$(6)$~\cite{wondrade2014gis}
}
\label{tab:datasets_citations}
\end{table}

%  'Benchmarking of wildland fire colour segmentation algorithms': 38, 'Traffic Signs Detection for Real-World Application of an Advanced Driving Assisting System Using Deep Learning': 46, '11K Hands: Gender recognition and biometric identification using a large dataset of hand images': 57, 'Determination of the age of oil palm from crown projection area detected from WorldView-2 multispectral remote sensing data: The case of Ejisu-Juaben district, Ghana': 59, 'GIS based mapping of land cover changes utilizing multi-temporal remotely sensed image data in Lake Hawassa Watershed, Ethiopia': 59

%% file: tables/funding_agencies.tex
\begin{table}[t]
\centering
\caption{Funding agencies and programs supporting African research and dataset acquisition}
\label{tab:funding_agencies}
\resizebox{\textwidth}{!}{
\begin{tabular}{|p{7cm}|p{8cm}|p{6cm}|p{5cm}|}
\hline
\textbf{Funding Agency/Program} & \textbf{Website URL} \\
\hline
AfriLabs & \url{https://www.afrilabs.com} \\
Code for Africa & \url{https://opportunities.codeforafrica.org} \\
The Engine Room & \url{https://www.theengineroom.org} \\
The Africa Data Hub (ADH) & \url{https://www.africadatahub.org} \\
African Union Development Agency (AUDA-NEPAD) & \url{https://www.nepad.org} \\
African Development Bank (AfDB) & \url{https://www.afdb.org} \\
%African Union & Promotes unity and solidarity of African countries & Broad African dataset acquisition & \url{https://au.int} \\
%\hline
Open Data Portal & \url{https://dataportals.org/about} \\
Lacuna Fund & \url{https://lacunafund.org/datasets} \\
Zindi & \url{https://zindi.africa} \\
Partnership for African Social and Governance Research (PASGR) & \url{https://www.pasgr.org} \\
DataFirst & \url{https://www.datafirst.com} \\
DS-I Africa Program (NIH, United States) & \url{https://www.nih.gov} \\
%DELTAS Africa (Wellcome Trust, United Kingdom) & Supports health research capacity & Sub-Saharan Africa & \url{https://wellcome.org} \\
%\hline
%Medical Research Council (MRC, United Kingdom) & Funds medical research & Sub-Saharan Africa & \url{https://mrc.ukri.org} \\
%\hline
%Bill and Melinda Gates Foundation (United States) & Funds global health and development projects & Sub-Saharan Africa & %\url{https://www.gatesfoundation.org} \\
%\hline
Horizon Europe (European Commission) & \url{https://ec.europa.eu/programmes/horizon2020} \\
%\hline
%Chan Zuckerberg Initiative & Supports science and education & Broad African research support & \url{https://chanzuckerberg.com} \\
%\hline
%Sloan Foundation & Funds research and education & Broad African research support & \url{https://sloan.org} \\
\hline
\end{tabular}}
\end{table}

%% file: content/topics.tex
\label{sec:topics}
\input{graphs/topics_taxonomy}
In this section, we study the research topics and recurring keywords in the computer vision field in Africa on the \emph{full} set of publications. In this analysis, we preferably use the \emph{full} set instead of the \emph{refined} set, since the \emph{refined} set depends on the top 50 keywords from global computer vision in their query generation. Thus, it can skew the results. These keywords are retrieved using an off-the-shelf tool. Out of 187,812 keywords, we only identify the top-30 recurring ones. We remove four keywords from this top-30 list corresponding to general computer vision research, which are of less interest in our fine-grained topics analysis, these keywords are (\emph{Computer Vision}, \emph{Camera}, \emph{Convolutional Neural Networks} and \emph{Deep Learning}). We manually categorize these keywords to create a taxonomy of the researched computer vision topics in the continent.

\input{graphs/keywords_region_notrotated}

Surprisingly, we found keywords such as \emph{Galaxies} and \emph{Crystalline Texture}, to verify their relevance we inspect five random publications for each. We found that \emph{Crystalline Texture} is used in publications related to texture analysis and classification which is relevant to computer vision. However, the keyword \emph{Galaxies} is used in publications that are relevant~(\shortciteA{fielding2022classification}), but others are not~(\shortciteA{shirley2021help}). In Figure~\ref{fig:topics_tax} we show the created taxonomy of the topic categories and the referred keywords. Three topics emerged during our keywords analysis that were not present when conducting the datasets analysis, which are \emph{Galaxies Morphology related}, \emph{Texture Analysis} and \emph{Image Encryption}. In Figure~\ref{fig:plot_highly_cited} (A-E), we show the distribution of these top-30 keywords per African region, to identify which region is the most contributing to that topic. Note that in Figure~\ref{fig:plot_highly_cited} (A) the term \emph{object detection} has around 70\% of the publications from Northern Africa, while the remaining ones are from sub-saharan Africa. Additionally, some papers can be counted multiple times as they include authors from different African regions.  

Inspecting the top-30 keywords distribution per keyword and African region, Figure~\ref{fig:plot_highly_cited} (A) shows that for \emph{Image Segmentation} Northern Africa contributes higher than other regions with around 90\%. Interestingly, Figure~\ref{fig:plot_highly_cited} (B) shows that \emph{Galaxies} is mostly researched in Southern Africa, which is hard to discern automatically whether it is relevant to computer vision or not as detailed earlier. Yet we identify some publications that are categorized under computer vision and it might be related to the Square Kilometre Array project that is hosted in Southern Africa. 
%The next common keywords are \emph{Animals}, \emph{Crops}, and keywords related to \emph{Remote Sensing} (e.g., \emph{Landsat}). 
Figure~\ref{fig:plot_highly_cited} (C, D) show both Eastern and Western Africa with keywords \emph{Landsat} and \emph{Land Cover} showing as around the second or third regions researching that topic, where Northern and Southern regions are mostly dominating these. %Figure~\ref{fig:plot_highly_cited} (E) shows that the most common keywords in Central Africa are related to \emph{Image Encryption}. 
The question of whether certain regions are working on their most urgent needs or not remains unanswered but is beyond the scope of this study. Nonetheless, this distribution of topics per region is an enabler for researchers and policy makers to make informed decisions on this previous question.

Finally, we compute the citations of the collected refined set publications per region. The top-1 cited paper per region are: (i) North Africa (\shortciteA{medhat2014sentiment}), (ii) Southern Africa (\shortciteA{van2014scikit}), (iii) Eastern Africa (\shortciteA{hengl2014soilgrids1km}), (iv) Western Africa (\shortciteA{nweke2018deep}) and (v) Central Africa (\shortciteA{potapov2012quantifying}).

%. Although, answering such a question is beyond this study but documenting the current state of computer vision research in Africa is the first step.
%\begin{itemize}
%\item a visualization of topic clusters and topic names overall
%\item Analysis per African region on the topics of interest
%\end{itemize}

%% file: graphs/topics_taxonomy.tex
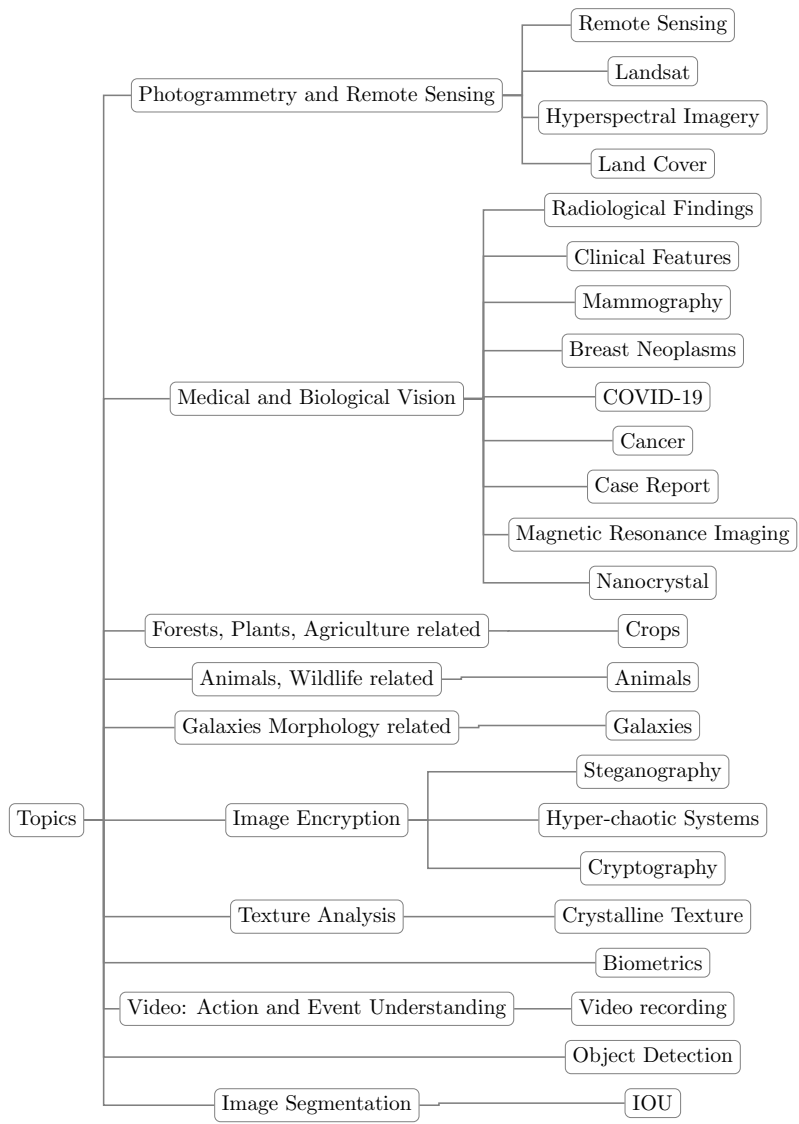
\begin{figure*}[!htbp]
    
\resizebox{0.7\textwidth}{!}{
\centering
\begin{forest} for tree={%
    draw=gray, minimum width=1cm, minimum height=.5cm, rounded corners=3,
    l sep+=5pt,
    grow'=east,
    edge={gray, thick},
    parent anchor=east,
    child anchor=west,
    if n children=0{tier=last}{},
    edge path={
      \noexpand\path [draw, \forestoption{edge}] (!u.parent anchor) -- +(10pt,0) |- (.child anchor)\forestoption{edge label};
    },
    if={isodd(n_children())}{
      for children={
        if={equal(n,(n_children("!u")+1)/2)}{calign with current}{}
      }
    }{}
    }
[Topics
    [Photogrammetry and Remote Sensing
        [Remote Sensing]
        [Landsat]
        [Hyperspectral Imagery]
        [Land Cover]
    ]
    [Medical and Biological Vision
        [Radiological Findings]
        [Clinical Features]
        [Mammography]
        [Breast Neoplasms]
        [COVID-19]
        [Cancer]
        [Case Report]
        [Magnetic Resonance Imaging]
        [Nanocrystal]
    ]
    [{Forests, Plants, Agriculture related}
        [Crops]
    ]
    [{Animals, Wildlife related}
        [Animals]]
    [Galaxies Morphology related
        [Galaxies]]
    [Image Encryption
        [Steganography] 
        [Hyper-chaotic Systems]
        [Cryptography]
    ]
    [Texture Analysis
        [Crystalline Texture]
    ]
    [Biometrics]
    [Video: Action and Event Understanding
        [Video recording]
    ]
    [Object Detection]
    [Image Segmentation
        [IOU]
    ]
]
\end{forest}
}
\caption{Topic categories and keywords taxonomies of the retrieved publications. The first level is the topic category, while the second level shows the keywords that are categorized under that topic.}
\label{fig:topics_tax}
\end{figure*}

%% file: graphs/keywords_region_notrotated.tex
\pgfplotstableread{ % data 
Label	FALSE	TRUE
Steganography	11.4	88.6
{Crystalline Texture}	12.1	87.9
Crops	59.5	40.5
Landsat	62.3	37.7
{Video recording}	14.8	85.2
{Hyperspectral Imagery}	19	81
Nanocrystal	34.3	65.7
{Radiological Findings}	21	79
{Clinical Features}	21	79
Animals	49.5	50.5
Mammography	23.1	76.9
{Breast Neoplasms}	25.9	74.1
Galaxies	97.1	2.9
Biometrics	20.6	79.4
{Land Cover}	62.9	37.1
{Image Segmentation}	8.9	91.1
COVID-19	25.9	74.1
{Hyper-chaotic Systems}	14.3	85.7
Cryptography	12.3	87.7
{Image Encryption}	11.6	88.4
{Remote Sensing}	49.5	50.5
Cancer	24	76
{Case Report}	25.8	74.2
{MRI}	22.2	77.8
IOU	28.3	71.7
{Object Detection}	27.6	72.4
}\testdataN

\pgfplotstableread{ % data 
Label	FALSE	TRUE
1	97.7	2.3
2 	94.5	5.5
3 	67.6	32.4
4 	73.8	26.2
5 	92.7	7.3
6 	87.8	12.2
7 	71.2	28.8
8 	92.5	7.5
9 	92.5	7.5
10 	68.1	31.9
11 	92.8	7.2
12 	88.9	11.1
13 	5.1	94.9
14 	88.3	11.7
15 	75.5	24.5
16 	96.1	3.9
17 	89.1	10.9
18 	98.7	1.3
19 	98.9	1.1
20 	98.7	1.3
21 	77.4	22.6
22 	88.2	11.8
23 	90	10
24 	87.2	12.8
25 	87.3	12.7
26 	87.5	12.5
}\testdataS

\pgfplotstableread{ % data  
Label	FALSE	TRUE
1	95.3	4.7
2	97.5	2.5
3	82.2	17.8
4	79.8	20.2
5	96.9	3.1
6	95.7	4.3
7	95.2	4.8
8	95	5
9	95	5
10	90.6	9.4
11	92.6	7.4
12	92.6	7.4
13	97.1	2.9
14	97.9	2.1
15	79.6	20.4
16	97.1	2.9
17	93.5	6.5
18	97.8	2.2
19	98.5	1.5
20	98.5	1.5
21	86.2	13.8
22	92.6	7.4
23	92.5	7.5
24	96	4
25	92.1	7.9
26	92.2	7.8
}\testdataE

\pgfplotstableread{ % data  
Label	FALSE	TRUE
1	95.3	4.7
2	95.9	4.1
3	86.5	13.5
4	81.2	18.8
5	95.6	4.4
6	97.2	2.8
7	93.4	6.6
8	89.2	10.8
9	89.2	10.8
10	91.6	8.4
11	89.1	10.9
12	91.5	8.5
13	98.7	1.3
14	91.6	8.4
15	80	20
16	97.6	2.4
17	88.8	11.2
18	96.8	3.2
19	96.4	3.6
20	97.1	2.9
21	85.1	14.9
22	92.8	7.2
23	91.5	8.5
24	94.1	5.9
25	91.3	8.7
26	91.7	8.3
}\testdataW

\pgfplotstableread{ % data 
Label	FALSE	TRUE
1	99.7	0.3
2	99.7	0.3
3	98.9	1.1
4	94.5	5.5
5	99.7	0.3
6	99.5	0.5
7	99.7	0.3
8	98.5	1.5
9	98.5	1.5
10	97.5	2.5
11	99.3	0.7
12	100	0.0
13	100	0.0
14	99.8	0.2
15	94.1	5.9
16	99.8	0.2
17	98.3	1.7
18	89.9	10.1
19	91.9	8.1
20	92.1	7.9
21	96	4
22	99.2	0.8
23	98.7	1.3
24	99.2	0.8
25	99.3	0.7
26	99.4	0.6
}\testdataC

%\begin{sidewaystable*}[!htbp]
%\begin{landscape}[!htbp]
\begin{figure*}[t]
%\begin{sidewaysfigure}
    \begin{subfigure}[b]{\linewidth}
        \centering
        \resizebox{\linewidth}{!}{
            \begin{tikzpicture}
                 \begin{axis}[
                    name=plot1,
                    xbar stacked,   % Stacked horizontal bars
                    bar width = 4pt,
                    xmin=0,         % Start x axis at 0
                    xmax=100,
                    ytick=data,     % Use as many tick labels as y coordinates
                    legend style={at={(0.5, 1.05)}, anchor=north, legend columns=-1},
                    yticklabels from table={\testdataN}{Label},  % Get the labels from the Label column of the \datatable
                    x label style={ at={(0.5, 1.23)},anchor=north, below=5mm},
                    xlabel = {(A) Northern Africa},
                    y tick label style={font=\tiny}
                    ]
                    \addplot table [x=FALSE, meta=Label,y expr=\coordindex] {\testdataN};   % "First" column against the data index
                    \addplot table [x=TRUE, meta=Label,y expr=\coordindex] {\testdataN};    
                    \legend{False, True}
                \end{axis}
           
               \begin{axis}[
                    name=plot2, at={(plot1.south east)},
                    %height=\textwidth,
                    %width=0.8\textwidth,
                    xbar stacked,   % Stacked horizontal bars
                    bar width = 4pt,
                    xmin=0,         % Start x axis at 0
                    xmax=100,
                    ytick=data,     % Use as many tick labels as y coordinates
                    legend style={at={(0.5, 1.05)}, anchor=north, legend columns=-1},
                    yticklabels=\empty,  % Get the labels from the Label column of the \datatable
                    x label style={at={(0.5, 1.23)},anchor=north, below=5mm},
                    xlabel = {(B) Southern Africa},
                    y tick label style={font=\tiny}
                    ]
                    \addplot table [x=FALSE, meta=Label,y expr=\coordindex] {\testdataS};   % "First" column against the data index
                    \addplot table [x=TRUE, meta=Label,y expr=\coordindex] {\testdataS};    
                    \legend{False, True}
                \end{axis}
   
                \begin{axis}[
                    name=plot3, at={(plot2.south east)},
                    %height=\textwidth,
                    %width=0.8\textwidth,
                    xbar stacked,   % Stacked horizontal bars
                    bar width = 4pt,
                    xmin=0,         % Start x axis at 0
                    xmax=100,
                    ytick=data,     % Use as many tick labels as y coordinates
                    legend style={at={(0.5, 1.05)}, anchor=north, legend columns=-1},
                    yticklabels=\empty, % Get the labels from the Label column of the \datatable
                    x label style={at={(0.5, 1.23)},anchor=north, below=5mm},
                    xlabel = {(C) Eastern Africa},
                    y tick label style={font=\tiny}
                    ]
                    \addplot table [x=FALSE, meta=Label,y expr=\coordindex] {\testdataE};   % "First" column against the data index
                    \addplot table [x=TRUE, meta=Label,y expr=\coordindex] {\testdataE};    
                    \legend{False, True}
                \end{axis}
            \end{tikzpicture}} 
    \end{subfigure}

    \begin{subfigure}[b]{\linewidth}
    \centering
    \resizebox{0.7\linewidth}{!}{
        \begin{tikzpicture}
                \begin{axis}[
                    name=plot4,
                    xbar stacked,   % Stacked horizontal bars
                    bar width = 4pt,
                    xmin=0,         % Start x axis at 0
                    xmax=100,
                    ytick=data,     % Use as many tick labels as y coordinates
                    legend style={at={(0.5, 1.05)}, anchor=north, legend columns=-1},
                    yticklabels from table={\testdataN}{Label},  % Get the labels from the Label column of the \datatable
                    x label style={at={(0.5, 1.23)},anchor=north, below=5mm},
                    xlabel = {(D) Western Africa},
                    y tick label style={font=\tiny}
                    ]
                    \addplot table [x=FALSE, meta=Label,y expr=\coordindex] {\testdataW};   % "First" column against the data index
                    \addplot table [x=TRUE, meta=Label,y expr=\coordindex] {\testdataW};    
                    \legend{False, True}
                \end{axis}

                \begin{axis}[
                    name=plot5, at={(plot4.south east)},
                    xbar stacked,   % Stacked horizontal bars
                    bar width = 4pt,
                    xmin=0,         % Start x axis at 0
                    xmax=100,
                    ytick=data,     % Use as many tick labels as y coordinates
                    legend style={at={(0.5, 1.05)}, anchor=north, legend columns=-1},
                    yticklabels=\empty,  % Get the labels from the Label column of the \datatable
                    x label style={at={(0.5, 1.23)},anchor=north, below=5mm},
                    xlabel = {(E) Central Africa},
                    y tick label style={font=\tiny}
                    ]
                    \addplot table [x=FALSE, meta=Label,y expr=\coordindex] {\testdataC};   % "First" column against the data index
                    \addplot table [x=TRUE, meta=Label,y expr=\coordindex] {\testdataC};    
                    \legend{False, True}
                \end{axis}
        \end{tikzpicture}} 
    \end{subfigure}
\caption{Keywords analysis of the top-30 recurring keywords. (A-E) Distribution of the most recurring keywords per African region. Red indicates the percentage of publications within the corresponding region indexed with that keyword. We remove four words from the top-30 set corresponding to general topics in computer vision (i.e., Computer Vision, Camera, Deep Learning, Convolutional Neural Networks) to focus on fine-grained topics.}
\label{fig:plot_highly_cited}
%\end{sidewaystable*}
%\end{sidewaysfigure}
\end{figure*}
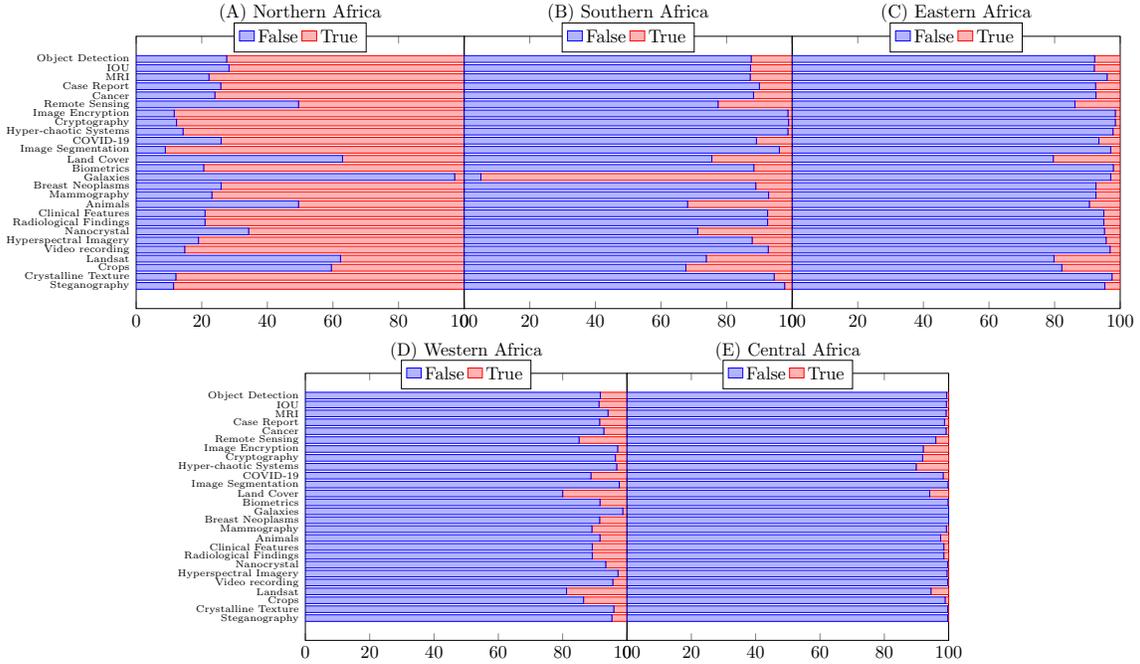
%\end{landscape}

%% file: content/publishing_trends.tex
%\subsection{Geo-temporal analysis}
%\textcolor{red}{Please move the data in the table authors tex file to the GitHub repository and we left it out for space.}
\input{graphs/refined_african_pubs}
\input{graphs/toptier_geotemporal}

We demonstrate the publishing patterns for the different African regions in the computer vision field over the last ten years (2012-2022). Figure~\ref{fig:geotemporal_refined} shows the number of publications, on the logarithmic scale, from the \emph{refined} set with at least one author working in an African institution. It shows that institutions in Northern and Southern Africa are the two highest regions publishing in computer vision, constructing around 88.5\% of the total publications. Looking at Eastern and Western Africa, we notice consistent growth of the publications over the period 2016-2022.

\input{graphs/collabs_africavsintern}
We perform a separate geo-temporal analysis on top-tier publications only. Figure~\ref{fig:toptier} shows the number of publication-researcher pairs in the aforementioned venues across the different continents over the last ten years. We show the publication-researcher pairs as we want to fairly document the number of researchers from African institutions that can publish in these leading venues. It clearly shows that North America and Asia are the ones mostly publishing in top-tier venues with around three quarters (74\%) of these publications stemming from them. However, when looking to Africa it only constitutes 0.06\% of the total publication-researcher pairs, and over the years it does not show a consistent increase and growth. Note, that for this analysis we only use the top-tier set of 45,000 publications and count the publication-researcher pairs as an indication of the computer vision capacity per continent. We provide additional analysis of the provided numbers per year and continent in our code repository\footnoteref{code}.
%Interestingly, we found that Madagascar had two researchers that published in \emph{ICLR} 2022, which shows the promise of how a small under-resourced country was still able to publish in a leading venue as such. 

To understand the collaboration patterns in African computer vision publications, we analyze the African \textit{vs.} international collaborations in terms of the number of publications over the last ten years. Figure~\ref{fig:collabs_africa_vs_international} demonstrates that most of the publications are dominated by international collaborations with a very minimal amount of African collaborations, forming only 3.9\% of the total publications. We believe that encouraging collaborations among African researchers can strengthen the continent's research eco-system, due to the fact that African countries share some of the problems and bottlenecks they are facing. Additionally, strengthening the African continent to have less dependency on developed countries can lead to its sovereignty on the data and the accompanied algorithms to improve their economies. We also show wide consensus among African researchers that such local collaborations is one of the top directions to improve the research eco-system in Africa in the following section.

%~\cite{pouris2014research,turki2023machine}

%% file: graphs/refined_african_pubs.tex
\begin{figure}
\centering
\resizebox{\textwidth}{!}{
\begin{tikzpicture}
\begin{axis} [
    width=13cm,
    height=5cm,
    bar width = 2pt,
    ybar = .05cm,
    ymode=log,
    xmin = 2011,
    xmax = 2023,
    xtick={2012, 2013,..., 2022},
    xticklabels={2012, 2013,..., 2022},
    % title = \textbf{},
    ylabel = Number of Publications,
    xlabel = Year,
    enlarge y limits = {abs=0.8},
    legend style={
			at={(0.5,-0.35)},
			anchor=north,
			legend columns=-1},
]

\addplot[red,fill=red] coordinates { (2022, 1379) (2021, 1212) (2020, 1057) (2019, 907) (2018, 808) (2017, 655) (2016, 626) (2015, 421) (2014, 429) (2013, 310) (2012, 300)};

\addplot[red!60,fill=red!60] coordinates { (2022, 188) (2021, 185) (2020, 179) (2019, 161) (2018, 111) (2017, 90) (2016, 59) (2015, 61) (2014, 49) (2013, 53) (2012, 39)};

\addplot[blue,fill=blue] coordinates { (2022, 173) (2021, 121) (2020, 91) (2019, 73) (2018, 40) (2017, 27) (2016, 21) (2015, 10) (2014, 19) (2013, 6) (2012, 13)};

\addplot[blue!60,fill=blue!60] coordinates { (2022, 221) (2021, 122) (2020, 67) (2019, 53) (2018, 40) (2017, 24) (2016, 16) (2015, 25) (2014, 13) (2013, 6) (2012, 7)};

\addplot[blue!30,fill=blue!30] coordinates { (2022, 17) (2021, 23) (2020, 16) (2019, 11) (2018, 9) (2017, 1) (2016, 2) (2015, 3) (2014, 3) (2013, 5) (2012, 1)};

\legend{Northern Africa, Southern Africa, Western Africa, Eastern Africa, Central Africa}

\end{axis}
\end{tikzpicture}
}
\caption{Scopus-indexed computer vision publications per African region across the time interval 2012-2022 showing the number of publications. We use the logarithmic scale. It shows consistent growth in Northern and Southern regions’ publications, and a recent increase in Eastern and Western Africa (2016-2022). However, Central Africa is the most in need of improving the computer vision capacity.}
\label{fig:geotemporal_refined}
\vspace{-1em}
\end{figure}
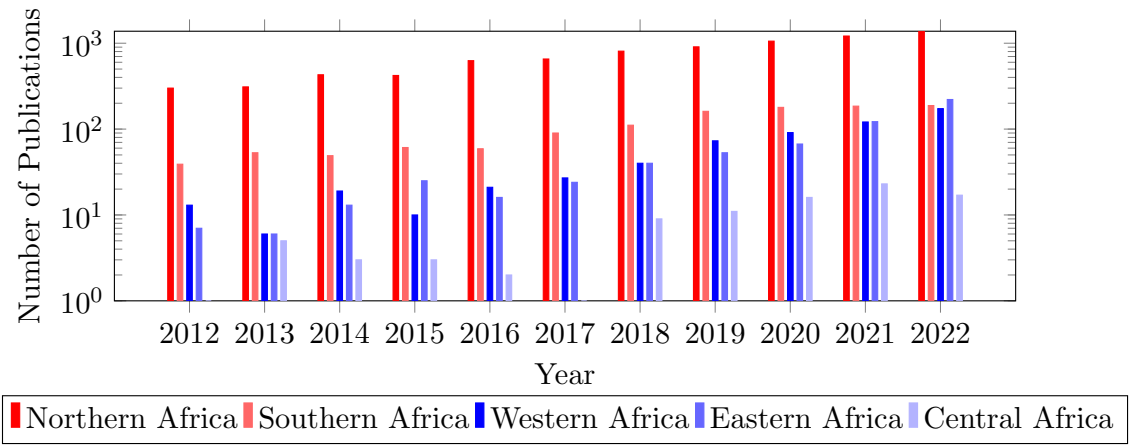

%% file: graphs/toptier_geotemporal.tex
\begin{figure}
    \centering
    \resizebox{\textwidth}{!}{
\begin{tikzpicture}
\begin{axis} [
    width=13cm,
    height=5cm,
    bar width = 2.5pt,
    ybar = .05cm,
    ymode=log,
    xmin = 2011,
    xmax = 2023,
    xtick={2012, 2013,..., 2022},
    % title = \textbf{},
    ylabel = Publication-Researcher Pairs,
    xlabel = Year,
    enlarge y limits = {abs=0.8},
    legend style={
			at={(0.5,-0.35)},
			anchor=north,
			legend columns=-1},
]

\addplot[black,fill=black] coordinates { (2022, 7498) (2021, 9926) (2020, 11073) (2019, 9120) (2018, 8620) (2017, 6058) (2016, 4487) (2015, 3813) (2014, 3486) (2013, 3461) (2012, 3030)};

\addplot[black!80,fill=black!80] coordinates { (2022, 14853) (2021, 11948) (2020, 10431) (2019, 7622) (2018, 4925) (2017, 3205) (2016, 2276) (2015, 2183) (2014, 1585) (2013, 1713) (2012, 1402)};

\addplot[black!60,fill=black!60] coordinates { (2022, 4892) (2021, 6060) (2020, 5962) (2019, 4860) (2018, 4318) (2017, 3319) (2016, 2432) (2015, 2358) (2014, 2006) (2013, 2297) (2012, 2095)};

\addplot[black!40,fill=black!40] coordinates { (2022, 1024) (2021, 803) (2020, 796) (2019, 611) (2018, 491) (2017, 378) (2016, 299) (2015, 284) (2014, 260) (2013, 267) (2012, 164)};

\addplot[black!20,fill=black!20] coordinates { (2022, 40) (2021, 83) (2020, 69) (2019, 32) (2018, 48) (2017, 21) (2016, 21) (2015, 39) (2014, 21) (2013, 27) (2012, 13)};

\addplot[red,fill=red] coordinates { (2022, 21) (2021, 9) (2020, 38) (2019, 8) (2018, 11) (2017, 3) (2016, 5) (2015, 2) (2014, 9) (2013, 2) (2012, 7)};

\legend{North America, Asia, Europe, Oceania, South America, Africa}

\end{axis}
\end{tikzpicture}
}
\caption{Publications in top-tier venues (\emph{CVPR, ICCV, ECCV, ICML, NeurIPS, ICLR, MICCAI, TPAMI, IJCV}) across all continents showing number of researcher-publication pairs per continent over the last ten years, with Africa highlighted in red.}
\label{fig:toptier}
\vspace{-1em}
\end{figure}
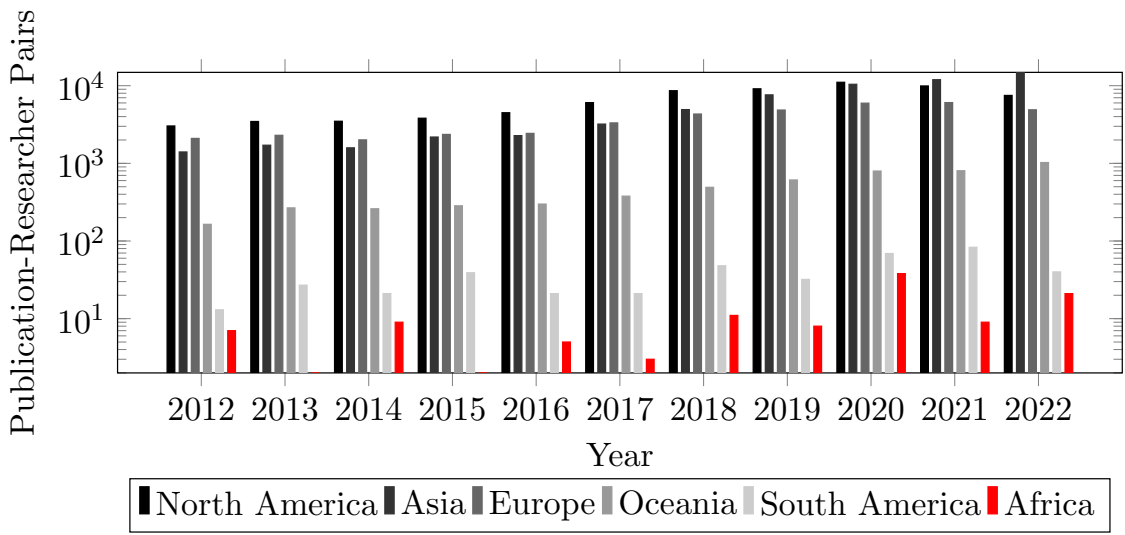

%% file: graphs/collabs_africavsintern.tex
\begin{figure}[t]
\centering
\resizebox{\textwidth}{!}{
\begin{tikzpicture}
\begin{axis} [
    width=13cm,
    height=5cm,
    bar width = 4pt,
    ybar = .05cm,
    ymode=log,
    xmin = 2011,
    xmax = 2023,
    xtick={2012, 2013,..., 2022},
    xticklabels={2012, 2013,..., 2022},
    % title = \textbf{},
    ylabel = Number of Publications,
    xlabel = Year,
    enlarge y limits = {abs=0.8},
    legend style={
			at={(0.5,-0.35)},
			anchor=north,
			legend columns=-1},
]

\addplot[black!50,fill=black!50] coordinates { (2022, 1809) (2021, 1128) (2020, 891) (2019,656) (2018, 519) (2017, 332) (2016, 329) (2015, 256) (2014, 267) (2013, 164) (2012, 181)};

\addplot[red,fill=red] coordinates { (2022, 88) (2021, 57) (2020, 45) (2019, 24) (2018, 25) (2017, 9) (2016, 8) (2015, 9) (2014, 7) (2013, 2) (2012, 5)};

\legend{International, African}

\end{axis}
\end{tikzpicture}
}
\caption{Collaboration patterns analysis showing the distribution of African versus international collaborations over the last ten years.}
\label{fig:collabs_africa_vs_international}
\end{figure}
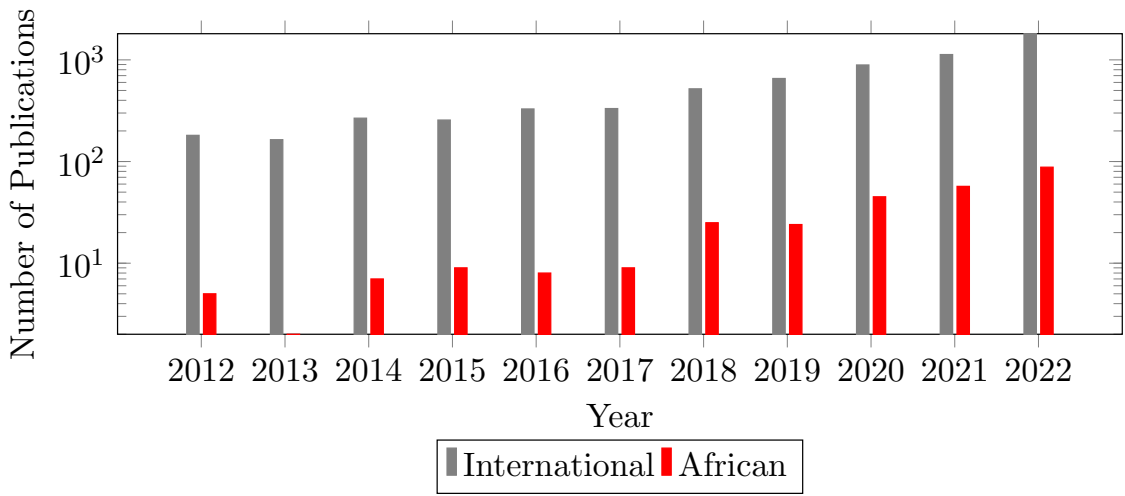

%% file: content/questionnaire.tex
\begin{figure}[t]
    \centering
    % \includesvg[scale=0.55]{graphs/citizen_residence}
    \includegraphics[scale=0.5]{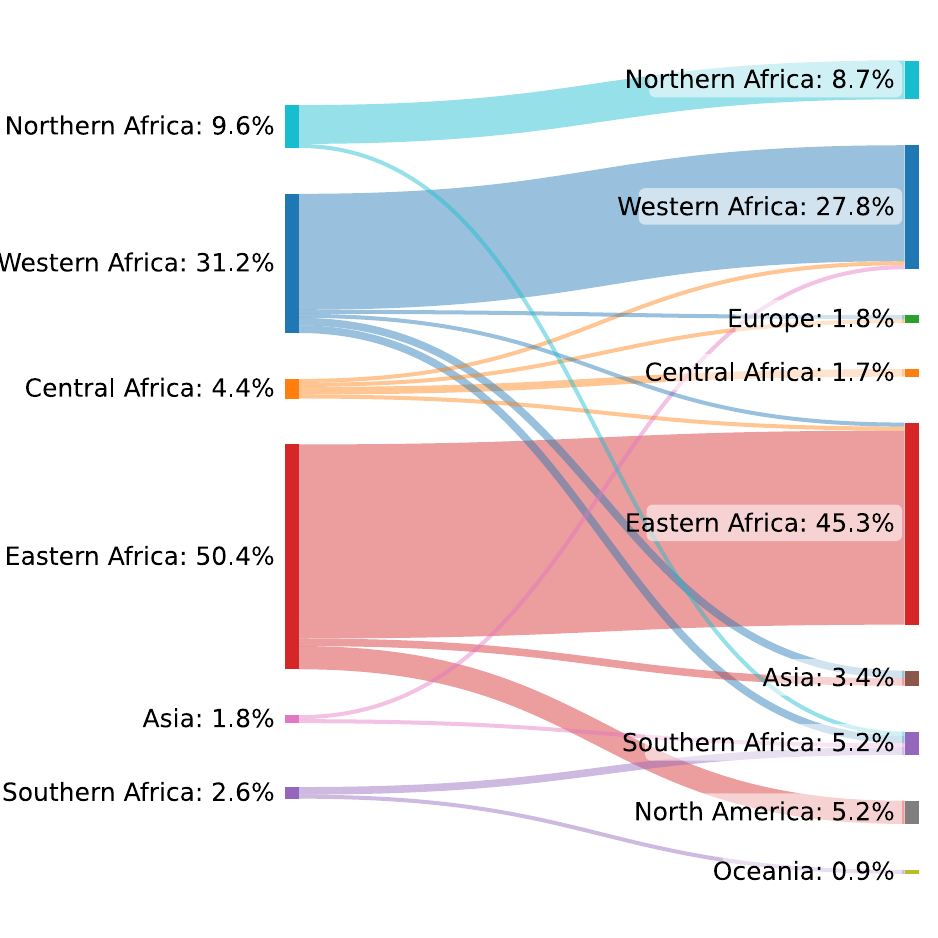}
    \caption{Participants' distribution in the region of citizenship (\emph{left}) against region of residence (\emph{right})}
    \label{fig:sankeymatic_chart}
\end{figure}

In this section, we document the results from our large-scale questionnaire to understand African researchers' view of the field and the barriers they face. We started with a pilot study \cite{omotayo2023towards}, where we discussed open questions within a participatory framework, and asked 14 community members from Egypt, Nigeria, Cameroon, and Benin about the barriers facing African computer vision researchers. The results from the pilot study were used to guide and refine the questions in the large-scale questionnaire. We publicly shared our large-scale survey through the Deep Learning Indaba platform to gain insights into the current issues from African researchers' views. We received 115 responses from researchers across the continent. Most participants of our questionnaire were graduate students, constituting 44\%, followed by undergraduate students at 28\%. Based on our analysis of the participants, we show the distribution of citizenship and residence of the participants in Figure~\ref{fig:sankeymatic_chart}. It can be observed that the majority of the participants are from Eastern Africa at 50.4\% followed by Western Africa at 31.2\% while the least participation is from the Southern region. It also demonstrates that the majority of the participants were not only African citizens, but they are also residing in Africa to better reflect the continent's research eco-system.

\input{graphs/questionnaire_Africa}

\input{graphs/questionnaire_position}
We start with an overview of the results followed by a fine-grained analysis and focus on two main questions:
(i) ``What can you identify as Top-3 setbacks/structural barriers in African computer vision research? Select from the list and/or add more under Others.''
(ii)``What do you think is the Top-2 urgent directions to improve the computer vision research eco-system in Africa?  Select from the list and/or add more under Others.'' Some of the choices for structural barriers include `lack of funding' and `poor economic systems'. The former results in the inability of research labs to conduct research, while the later entails the escape of talent abroad to avoid such economic situations and inability to build proper infrastructure necessary for research. For the first question we found the highest three barriers were lack of funding, low access to resources (e.g., compute) and limited availability of African datasets. For the second question we found the top two directions from the participants views were, establishing research collaborations across African universities and launching projects to curate African computer vision datasets.
%The main question we show here is about the barriers that face researchers conducting machine learning and computer vision research in the continent. 

For the fine-grained analysis we show the distribution of the responses per African region and per career position for the former two questions. Note that participants chose multiple answers for these questions and percentages are computed with respect to the total participants per answer. Figure \ref{fig:plot_questionnaire_global} shows the distribution with respect to the African regions for the two former questions. The responses from Eastern Africa agreed more on ``Barrier 1: Lack of funding'' for the first question and ``Direction 1: Better computer vision courses taught in universities'' for the second. While South Africans appear to agree more on ``Barrier 3: Low access to resources'' for the first question and ``Direction 4: Launching projects to curate African computer vision datasets'' for the second.

Figure~\ref{fig:questionnaire_pos} shows fine-grained analysis with respect to career position for both questions. Although graduate students agree more on ``Direction 1: Better computer vision courses taught in universities'' for the second question at 54.5\%, industry professionals hardly agree with this at 21.4\%. They rather see that directions towards establishing ``Direction 2: Collaborations among African universities'' and ``Direction 4: Launching projects to curate African datasets'' more important. The collected responses also support the recommendation by~\cite{omotayo2023towards} that establishing a balance of both internal and external collaboration is one of the important tools to address these barriers. We also believe the progressive spread of cutting-edge techniques in computer vision through local training and regional competitions, could promote technical expertise and ensure the availability of datasets. We also believe that establishing African collaborations on the university level and improving computer vision courses are important directions to pursue.
%\todo{ Conversely, it is observed that their is no widely agreed opinion \men{I disagree there is an obvious agreed upon opinion on that one, I would also recommend adding the percentages for each here and in the previous statement} regarding the top-4 urgent direction to improve CV research in Africa (i-Better computer vision courses taught in universities, ii-Research collaborations across African universities/research institutes, iii-International collaborations with improved access to large-scale compute and iv-Encourage and launch projects to curate African computer vision datasets). Additionally, there is wide acceptance of potential CV summer school  as well as the possibility to hold a 2-day CV curriculum retreat for university faculties.}
%\men{We can keep these as well and report again percentages but if you can remove them from the figures that would be great.}

% \todo{Charts in Tikz}

%\input{graphs/questionnaire_position}

% ***discussion section
%\luk{
%Provided results further highlights the differences in research interest across African researchers and the need to strengthen collaborations among African Universities and Institutions.  We therefore emphasize that the opportunity to improve how CV is thought in African educational institutions must receive all forms of support to ensure its sustained improvement.}

%% file: graphs/questionnaire_Africa.tex
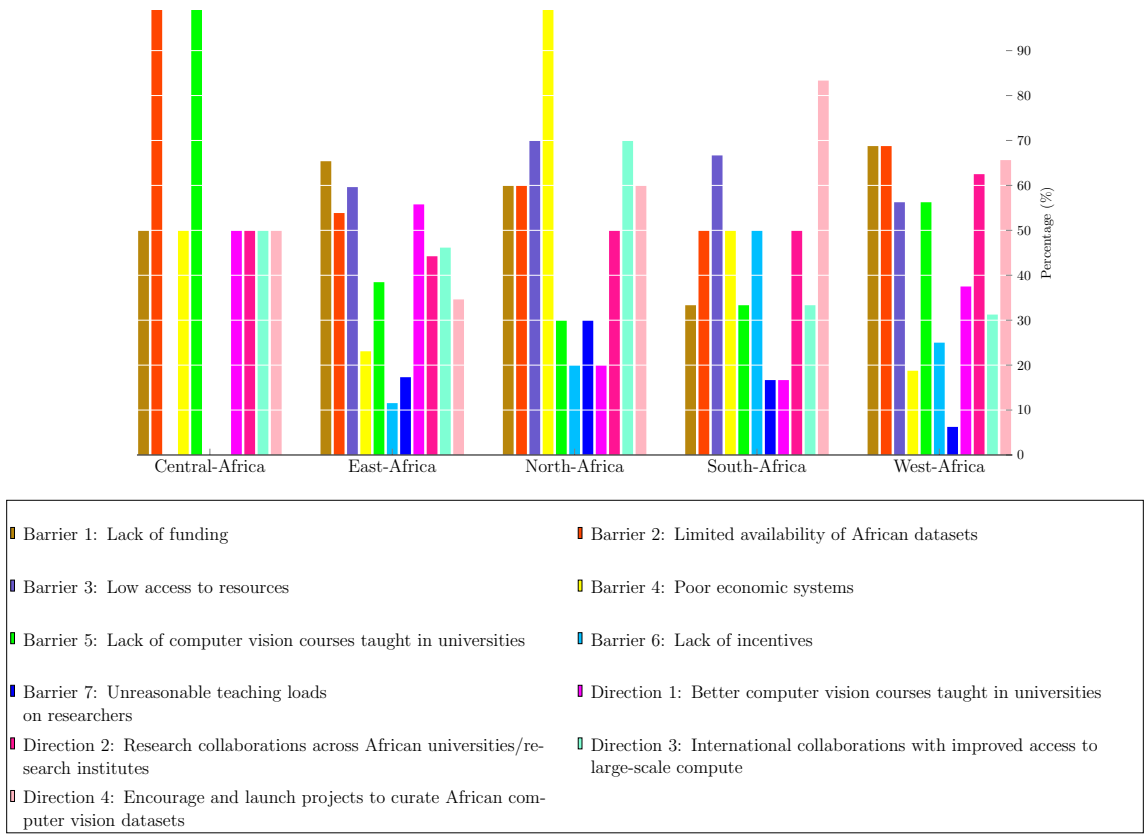
\begin{figure*}[t]
\begin{center}
\resizebox{\textwidth}{!}{
 \begin{tikzpicture} %[rotate=90] %thick
  \centering
  \begin{axis}[
        ybar, axis on top,
        title={},
        height=14cm, width=26cm,
        bar width=0.3cm,
        ymajorgrids, tick align=inside,
        major grid style={draw=white},
        enlarge y limits={value=.1, upper},
        ymin=0, ymax=90,
        axis x line*=bottom,
        axis y line*=right,
        y axis line style={opacity=0},
        tickwidth=5pt,
        enlarge x limits=true,
        legend style={
                font=\Large,
                at={(1.15,-0.85)},
                anchor=south east,
                column sep=1ex,
                row sep = -2ex,
                legend columns=2,
                minimum height=1.8cm, 
                text width=1\textwidth,
        },
        legend cell align=left,
        ylabel={Percentage (\%)},
        xticklabel style={font=\Large},
        symbolic x coords={
           Central-Africa, East-Africa, North-Africa, South-Africa, West-Africa},
       xtick=data,
       % nodes near coords={
       %  \pgfmathprintnumber[precision=0]{\pgfplotspointmeta}
       % }
    ]
   \addplot [draw=none, fill=q3] coordinates {
      (Central-Africa, 50.0)
      (East-Africa, 65.38) 
      (North-Africa, 60.0)
      (South-Africa, 33.33) 
      (West-Africa, 68.75)};  
    \addplot [draw=none, fill=q4] coordinates {
      (Central-Africa, 100.0)
      (East-Africa, 53.85) 
      (North-Africa, 60.0)
      (South-Africa, 50.0) 
      (West-Africa, 68.75)};  
   \addplot [draw=none,fill=q5] coordinates {
      (Central-Africa, 00.0)
      (East-Africa, 59.62) 
      (North-Africa, 70.0)
      (South-Africa, 66.67) 
      (West-Africa, 56.25)};  
   \addplot [draw=none, fill=q6] coordinates {
      (Central-Africa, 50.0)
      (East-Africa, 23.08) 
      (North-Africa, 100.0)
      (South-Africa, 50.0) 
      (West-Africa, 18.75)};  
    \addplot [draw=none, fill=q7] coordinates {
      (Central-Africa, 100.0)
      (East-Africa, 38.46) 
      (North-Africa, 30.0)
      (South-Africa, 33.33) 
      (West-Africa, 56.25)};  
   \addplot [draw=none,fill=q8] coordinates {
      (Central-Africa, 00.0)
      (East-Africa, 11.54) 
      (North-Africa, 20.0)
      (South-Africa, 50.0) 
      (West-Africa, 25.0)};  
   \addplot [draw=none, fill=q9] coordinates {
      (Central-Africa, 00.0)
      (East-Africa, 17.31) 
      (North-Africa, 30.0)
      (South-Africa, 16.67) 
      (West-Africa, 06.25)};  
    \addplot [draw=none, fill=q10] coordinates {
      (Central-Africa, 50.0)
      (East-Africa, 55.77) 
      (North-Africa, 20.0)
      (South-Africa, 16.67) 
      (West-Africa, 37.50)};  
   \addplot [draw=none,fill=q11] coordinates {
      (Central-Africa, 50.0)
      (East-Africa, 44.23) 
      (North-Africa, 50.0)
      (South-Africa, 50.0) 
      (West-Africa, 62.50)};  
   \addplot [draw=none, fill=q12] coordinates {
      (Central-Africa, 50.0)
      (East-Africa, 46.15) 
      (North-Africa, 70.0)
      (South-Africa, 33.33) 
      (West-Africa, 31.25)};  
   \addplot [draw=none, fill=q13] coordinates {
      (Central-Africa, 50.0)
      (East-Africa, 34.62) 
      (North-Africa, 60.0)
      (South-Africa, 83.33) 
      (West-Africa, 65.63)};  
      
    \legend{
Barrier 1: Lack of funding,
Barrier 2: Limited availability of African datasets,
Barrier 3: Low access to resources, 
Barrier 4: Poor economic systems,
Barrier 5: Lack of computer vision courses taught in universities,
Barrier 6: Lack of incentives,
Barrier 7: Unreasonable teaching loads\\ on researchers,
Direction 1: Better computer vision courses taught in universities,
Direction 2: Research collaborations across African universities/research institutes,
Direction 3: International collaborations with improved access to large-scale compute,
Direction 4: Encourage and launch projects to curate African computer vision datasets} 
  \end{axis}
  \end{tikzpicture}
}
\end{center}
\vspace{-0.5em}
\caption{Fine-grained analysis based on regional representation within Africa. It shows the answers for two questions on the top-3 barriers facing African researchers and the top-2 urgent directions they believe are important to tackle. Participants can choose multiple answers, and the percentage is the number of responses for a specific answer with respect to the total participation per region.}
\label{fig:plot_questionnaire_global}
\end{figure*}

%% file: graphs/questionnaire_position.tex
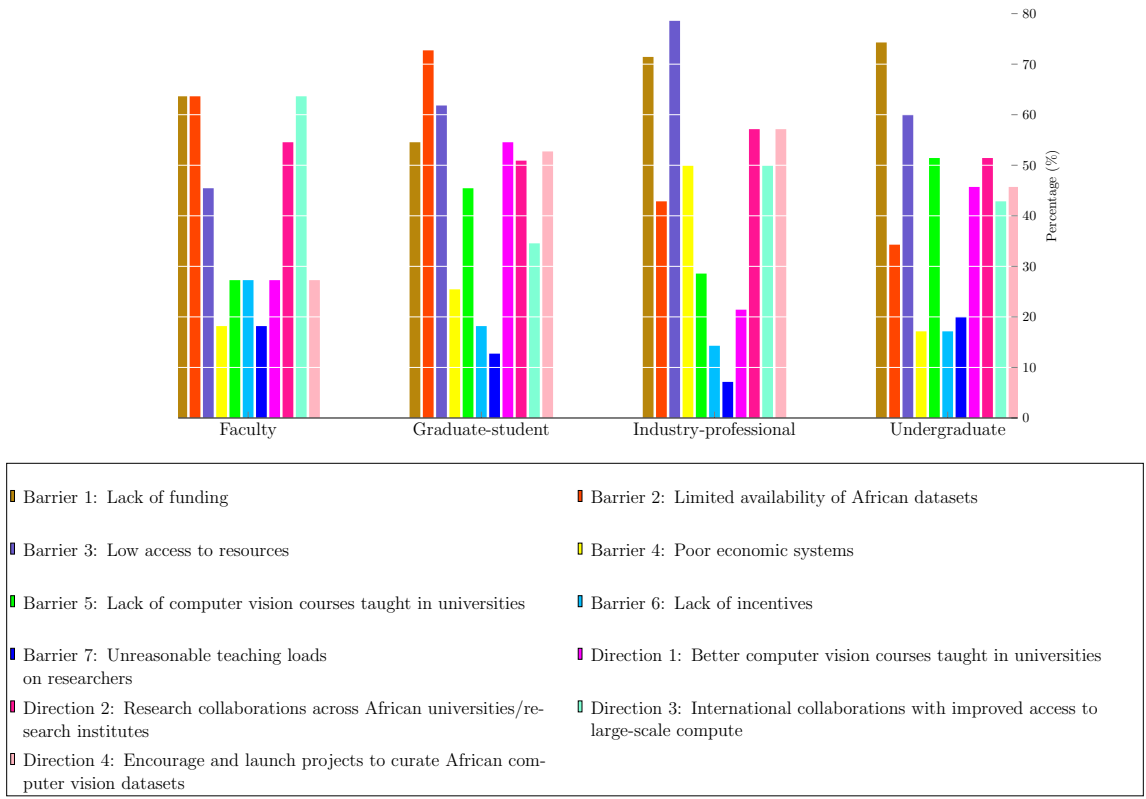
\begin{figure*}[!t]
\centering
\resizebox{\textwidth}{!}{
\begin{tikzpicture} [rotate=0] %thick
  \centering
  \begin{axis}[
        ybar, axis on top,
        height=14cm, width=25cm,
        bar width=0.3cm,
        ymajorgrids, tick align=inside,
        major grid style={draw=white},
        enlarge y limits={value=.1, upper},
        ymin=0, ymax=80,
        axis x line*=bottom,
        axis y line*=right,
        y axis line style={opacity=0},
        tickwidth=5pt,
        enlarge x limits=true,
        legend style={
                font=\Large,
                at={(1.15,-0.85)},
                anchor=south east,
                column sep=1ex,
                row sep = -2ex,
                legend columns=2,
                minimum height=1.8cm, 
                % nodes={scale=0.5, transform shape}
                text width=1\textwidth,
        },
        legend cell align=left,
        ylabel={Percentage (\%)},
        xticklabel style={font=\Large},
        symbolic x coords={
           Faculty, Graduate-student, Industry-professional, Undergraduate},
       xtick=data,
    ]
   \addplot [draw=none, fill=q3] coordinates {
      (Faculty, 63.64)
      (Graduate-student, 54.55)
      (Industry-professional, 71.43) 
      (Undergraduate, 74.29)};
    \addplot [draw=none, fill=q4] coordinates {
      (Faculty, 63.64)
      (Graduate-student, 72.73)
      (Industry-professional, 42.86) 
      (Undergraduate, 34.29)};
   \addplot [draw=none,fill=q5] coordinates {
      (Faculty, 45.45)
      (Graduate-student, 61.82)
      (Industry-professional, 78.57) 
      (Undergraduate, 60.0)};
   \addplot [draw=none, fill=q6] coordinates {
      (Faculty, 18.18)
      (Graduate-student, 25.45)
      (Industry-professional, 50.00) 
      (Undergraduate, 17.14)};
    \addplot [draw=none, fill=q7] coordinates {
      (Faculty, 27.27)
      (Graduate-student, 45.45)
      (Industry-professional, 28.57) 
      (Undergraduate, 51.43)};
   \addplot [draw=none,fill=q8] coordinates {
      (Faculty, 27.27)
      (Graduate-student, 18.18)
      (Industry-professional, 14.29) 
      (Undergraduate, 17.14)};
   \addplot [draw=none, fill=q9] coordinates {
      (Faculty, 18.18)
      (Graduate-student, 12.73)
      (Industry-professional, 07.14) 
      (Undergraduate, 20.0)};
    \addplot [draw=none, fill=q10] coordinates {
      (Faculty, 27.27)
      (Graduate-student, 54.55) 
      (Industry-professional, 21.43) 
      (Undergraduate, 45.71)};
   \addplot [draw=none,fill=q11] coordinates {
      (Faculty, 54.55)
      (Graduate-student, 50.91)
      (Industry-professional, 57.14) 
      (Undergraduate, 51.43)};
   \addplot [draw=none, fill=q12] coordinates {
      (Faculty, 63.64)
      (Graduate-student, 34.55)
      (Industry-professional, 50.00) 
      (Undergraduate, 42.86)};
   \addplot [draw=none, fill=q13] coordinates {
      (Faculty, 27.27)
      (Graduate-student, 52.73)
      (Industry-professional, 57.14) 
      (Undergraduate, 45.71)};

    \legend{Barrier 1: Lack of funding, Barrier 2: Limited availability of African datasets, 	Barrier 3: Low access to resources, 
Barrier 4: Poor economic systems,
Barrier 5: Lack of computer vision courses taught in universities,
Barrier 6: Lack of incentives,
Barrier 7: Unreasonable teaching loads\\ on researchers,
Direction 1: Better computer vision courses taught in universities,
Direction 2: Research collaborations across African universities/research institutes,
Direction 3: International collaborations with improved access to large-scale compute,
Direction 4: Encourage and launch projects to curate African computer vision datasets} 
  \end{axis}
  \end{tikzpicture} }
% 13 9
% \fbox{
\vspace{-0.5em}
\caption{Fine-grained Analysis based on career position. It shows the answers for two questions on the top-3 barriers facing African researchers and the top-2 urgent directions they believe are important to tackle. Participants can choose multiple answers, and the percentage is the number of responses for a specific answer with respect to the total participation per position.}
\label{fig:questionnaire_pos}
\end{figure*} 

%% file: content/conclusion.tex
We present a case study on African computer vision research and study the inequity within the continent and with respect to the global context in terms of publications. Furthermore, we provide taxonomies for the datasets and topics researched in the continent. Our study also provides a catalog of datasets to aid small-scale projects in Africa and to encourage and launch projects to curate computer vision datasets within the taxonomy of listed research topics. Moreover, we have shown per region distribution of the most recurring research topics in computer vision in the continent to guide researchers and policy makers in identifying whether African research aligns with the communities' needs or not. 
%Additionally, our study documented inequity in computer vision research among the different regions and globally. 
Finally, a large-scale questionnaire revealed consensus among participants on key barriers and emphasizing the urgent need for internal collaborations, as outlined in our study. For our future work, we aim to focus on the creation of an academic committee to discuss computer vision syllabi and its dissemination through courses or summer schools. Our community has contributed to the first African computer vision summer school (ACVSS)\footnote{\url{https://www.acvss.ai}}. Similar initiatives exist such as the RISE-MICCAI Winter and Summer Schools\footnote{\url{https://miccai.org/index.php/about-miccai/rise-miccai/}} and the ACM SIGIR/SIGKDD African Summer School on Machine Learning for Data Mining and Search\footnote{\url{https://sigir.org/afirm2020/}}. These provide computer vision and artificial intelligence training for African researchers with full scholarships or affordable registration fees. Additionally, we aim to encourage African projects that rely on our provided listing of datasets to build computer vision capacity in the continent.

\textbf{Limitations:} While our study offers valuable insights into the field of computer vision research, it is essential to recognize its limitations. One notable limitation is our reliance on Scopus as the primary data source. It is mostly tied to venues with high publishing costs. These costs can create barriers for researchers in lower-income regions, especially in Africa. Moreover, Scopus is mainly dominated by publications in the English language, where other languages used in Africa e.g., Arabic, Swahili or French would be missing. As a result, African researchers may publish more frequently in alternative venues not indexed by Scopus. Incorporating additional databases such as arXiv could provide a more comprehensive perspective. Another limitation is the potential for inaccurate retrieval of relevant literature due to the use of our query (``image'' OR ``computer vision''). Determining the recall of our query is challenging, and it is likely that our automatic analysis missed some pertinent papers or retrieved irrelevant ones. Exploring more sophisticated search strategies could improve the robustness and completeness of the analysis.

%Our investigation revealed a total of 96 officially published datasets and 33 unofficial datasets. We meticulously examined 40 categories with a focus on benefiting African communities, particularly emphasizing image classification, object detection, and diverse applications.

%Analyzing the publishing patterns in African computer vision from 2012 to 2022, we found that Northern and Southern Africa accounted for 88.5\% of publications, while Eastern and Western Africa showed a noteworthy growth trend from 2016 to 2022. Notably, when scrutinizing top-tier publications, North America and Asia dominated, with Africa contributing only 0.06\%. This underscores the critical need for heightened collaboration to strengthen the continent's research ecosystem.

%Our study identified the top 30 recurring keywords, their distribution across African regions, co-occurrence patterns, and noteworthy observations. 

%% file: content/acknowledgement.tex
We thank \emph{SisonkeBiotik}, the Machine Learning for Health Research in Africa, and \emph{Masakhane}, the African community for Natural Language Processing, for their support in the development of this research paper. All the authors are members of \emph{Ro'ya-CV4Africa}, a community dedicated to Computer Vision by Africans and for Africans.